\documentclass[a4paper,twocolumn]{article}

\usepackage{authblk}
\usepackage{times}
\usepackage{epsfig}
\usepackage{graphicx}
\usepackage{amsmath}
\usepackage{amssymb}
\usepackage{graphicx}
\usepackage{comment}
\usepackage{xcolor}
\usepackage{subfigure}
\usepackage{url}
\usepackage[utf8]{inputenc}
\usepackage{diagbox}
\usepackage{pifont}
\usepackage{multirow}
\usepackage{array}
\usepackage{pdfpages}

\makeatletter
\renewcommand\AB@affilsepx{\space\protect\Affilfont}
\makeatother

\newcolumntype{P}[1]{>{\centering\arraybackslash}p{#1}}
\newcolumntype{M}[1]{>{\centering\arraybackslash}m{#1}}

\usepackage{array}
\newcommand{\PreserveBackslash}[1]{\let\temp=\\#1\let\\=\temp}
\newcolumntype{C}[1]{>{\PreserveBackslash\centering}m{#1}}
\newcolumntype{R}[1]{>{\PreserveBackslash\raggedleft}m{#1}}
\newcolumntype{L}[1]{>{\PreserveBackslash\raggedright}m{#1}}
\usepackage[breaklinks=true,bookmarks=false]{hyperref}

\usepackage[a4paper,top=3cm,bottom=2cm,left=2.5cm,right=2.5cm,marginparwidth=1.75cm]{geometry}
\usepackage{multicol}
\usepackage{parskip}

\title{\Large\textbf{Context-Aware Personality Inference in Dyadic Scenarios:\\ Introducing the UDIVA Dataset}}

\author[1,2]{Cristina Palmero\thanks{These authors contributed equally to this work.}}
\author[1,2]{Javier Selva$^*$}
\author[1,2]{Sorina Smeureanu$^*$}
\author[2,3]{Julio C. S. Jacques Junior}
\author[1,2]{Albert Clapés}
\author[1]{Alexa Moseguí}
\author[1,2]{Zejian Zhang}
\author[1]{David Gallardo}
\author[1]{Georgina Guilera}
\author[1]{\linebreak David Leiva}
\author[1,2]{Sergio Escalera}

\affil[1]{Universitat de Barcelona}
\affil[2]{Computer Vision Center}
\affil[3]{Universitat Oberta de Catalunya}
\affil[ ]{\tt\small \{crpalmec7, ssmeursm28, zzhangzh45\}@alumnes.ub.edu, jaselvaca@ub.edu,}
\affil[ ]{\tt\small jsilveira@uoc.edu, aclapes@cvc.uab.cat, alexa.moseguis@gmail.com\\\linebreak}
\affil[ ]{\tt\small \{david.gallardo, gguilera, dleivaur\}@ub.edu, sergio@maia.ub.es}

\date{}

\begin{document}
\maketitle

\setlength{\parskip}{0.1in}
\setlength{\parindent}{15pt}

\begin{abstract}
This paper introduces UDIVA, a new non-acted dataset of face-to-face dyadic interactions, where interlocutors perform competitive and collaborative tasks with different behavior elicitation and cognitive workload. The dataset consists of 90.5 hours of dyadic interactions among 147 participants distributed in 188 sessions, recorded using multiple audiovisual and physiological sensors. Currently, it includes sociodemographic, self- and peer-reported personality, internal state, and relationship profiling from participants. As an initial analysis on UDIVA, we propose a transformer-based method for self-reported personality inference in dyadic scenarios, which uses audiovisual data and different sources of context from both interlocutors to regress a target person's personality traits. Preliminary results from an incremental study show consistent improvements when using all available context information.
\end{abstract}

\section{Introduction}
\label{sec:intro}
Human interaction has been a central topic in psychology and social sciences, aiming at explaining the complex underlying mechanisms of communication with respect to cognitive, affective, and behavioral perspectives~\cite{burgoon2007interpersonal, braithwaite2014engaging}. From a computational point of view, research in dyadic and small group interactions enables the development of automatic approaches for detection, understanding, modeling, and synthesis of individual and interpersonal social signals and dynamics~\cite{vinciarelli2011bridging}. Many human-centered applications for good (e.g., early diagnosis and intervention~\cite{el2006affective}, augmented telepresence~\cite{ahuja2019react}, and personalized agents~\cite{esposito2016modeling}) strongly depend on devising solutions for such tasks.

In dyadic interactions, we use verbal and nonverbal communication channels to convey our goals and intentions~\cite{narayanan2013behavioral, VINCIARELLI20091743} while building a common ground~\cite{clark1983common}. Both interlocutors influence each other based on the cues we perceive~\cite{burgoon2007interpersonal}. However, the way we perceive, interpret, react, and adapt to them depends on a myriad of factors. Such factors, which we refer to as context, may include, but are not limited to: our personal characteristics, either stable (e.g., personality~\cite{cuperman2009big}, cultural background, and other sociodemographic information~\cite{segerstrale2018nonverbal}) or transient (e.g., mood~\cite{clore2007emotions}, physiological or biological factors); the relationship and shared history between both interlocutors; the characteristics of the situation and task at hand; societal norms; and environmental factors (e.g., temperature). What is more, to analyze individual behaviors during a conversation, the joint modeling of both interlocutors is required due to the existing dyadic interdependencies. While these aspects are usually contemplated in non-computational dyadic research~\cite{kenny1996models}, context- and interlocutor-aware computational approaches are still scarce, largely due to the lack of datasets providing contextual metadata in different situations and populations~\cite{dudzik2019context}.

Here, we introduce UDIVA, a highly varied multimodal, multiview dataset of zero- and previous-acquaintance, face-to-face dyadic interactions. It consists of 188 interaction sessions, where 147 participants arranged in dyads performed a set of tasks in different circumstances in a lab setting. It has been collected using multiple audiovisual and physiological sensors, and currently includes sociodemographic, self- and peer-reported personality, internal state, and relationship profiling. To the best of our knowledge, there is no similar publicly available, face-to-face dyadic dataset in the research field in terms of number of views, participants, tasks, recorded sessions, and context labels.

As an initial analysis on the UDIVA dataset, we also propose a novel method for self-reported personality inference in dyadic scenarios. Apart from its importance in interaction understanding, personality recognition is key to develop individualized, empathic, intelligent systems~\cite{park2020looking}. Our proposal is based on the Video Action Transformer~\cite{Girdhar_2019_CVPR}, which classifies people’s actions in a video by taking advantage of the spatiotemporal context around them. Inspired by~\cite{rahman-etal-2020-integrating}, we extend query, key, and value from~\cite{Girdhar_2019_CVPR} with the other interlocutor’s scene, audio, and further context metadata. The latter includes stable and transient characteristics from each interlocutor, as well as session, task, and relationship information. Finally, we experimentally evaluate the usefulness of each additional input incrementally, showing consistent improvements when using all the available context sources and modalities.

\section{Related work}
\label{sec:related}
\begin{table*}[t!]
\centering
  \setlength\tabcolsep{2.8pt}
  \scriptsize
\caption{Publicly available audiovisual human-human (face-to-face) dyadic interaction datasets. ``Interaction'', \textit{Acted} (actors improvising and/or following an interaction protocol, i.e. given topics/stimulus/tasks), \textit{Acted}$^*$ (Scripted), \textit{Non-acted} (natural interactions in lab environment) or \textit{Non-acted}$^*$ (non-acted but guided by interaction protocol); ``F/M'', number of participants per gender (Female/Male) or number of participants if gender is not informed; ``Sess'', number of sessions; ``Size'', hours of recordings;``\#Views'', number of RGB cameras used, and \textit{D} is RGB+D, \textit{E} is Ego, \textit{M} is Monochrome. The $\phi$ symbol is used to indicate missing/incomplete/unclear information on the source. }

\resizebox{\textwidth}{!}{%
\begin{tabular}{|C{1.95cm}|C{2.3cm}|C{1.4cm}|C{2cm}|C{3.5cm}|C{0.65cm}|C{0.8cm}|C{0.8cm}|C{0.9cm}|C{1cm}|}
\hline
\textbf{Name / Year} 
& \textbf{Focus} 
& \textbf{Interaction} 
& \textbf{Modality} 
& \textbf{Annotations}
& \textbf{F/M}
& \textbf{Sess}
& \textbf{Size}
& \textbf{\#Views}
& \textbf{Lang.}\\ \hline

IEMOCAP \cite{Busso2008}, 2008
& Emotion recognition 
& Acted$^*$ \& Acted 
& Audiovisual, face \& hands MoCap.
& Emotions, transcripts, turn-taking
& 5/5
& 5
& $\sim$12h 
& 2
& English \\ \hline

CID \cite{blache2017corpus}, 2008
& Speech \& conversation analysis
& Non-acted \& Non-acted$^*$
& Audiovisual
& Speech features, transcripts
& 10/6
& 8
& 8h 
& 1
& French \\ \hline

HUMAINE$^{\dag}$ \cite{douglas2007humaine,douglas2011humaine}, 2011
& Emotion analysis
& Non-acted$^*$
& Audiovisual
& Emotions
& 34
& 18
& $\sim$12h 
& 4
& English \\ \hline

MMDB \cite{rehg2013decoding}, 2013
& Adult-infant interaction analysis
& Non-acted$^*$
& Audiovisual, depth, physiological
& Social cues (gaze, vocal affects, gestures...)
& 121
& 160
&$\sim$13.3h
& 8 + 1D
& English \\ \hline

MAHNOB \cite{bilakhia2015mahnob}, 2015
& Mimicry
& Non-acted$^*$
& Audiovisual, head MoCap.
& Head, face and hand gestures, personality scores (self-reported)
& 29/31
& 54
& 11.6h
& 2 + 13M
& English \\ \hline

MIT Interview \cite{naim2015automated}, 2015
& Hirability analysis 
& Non-acted$^*$
& Audiovisual
& Hirability, speech features, social \& behavioral traits, transcripts
& 43/26
& 138
& 10.5h
& 2
& English \\ \hline

Creative IT \cite{Metallinou:2016}, 2016
& Emotion recognition
& Acted
& Audiovisual, body MoCap.
& Transcripts, speech features, emotion
& 9/7
& 8
& $\sim$1h
& 2
& English \\ \hline
MSP-IMPROV \cite{Busso_2017}, 2017
& Emotion recognition
& Acted \& Non-acted
& Audiovisual
& Turn-taking, emotion
& 6/6
& 6
& 9h
& 2
& English \\ \hline

DAMI-P2C \cite{chen2020dyadic}, 2020
& Adult-infant interaction analysis
& Non-acted$^*$ 
& Audiovisual 
& Emotion, sociodemographics, parenting assessment, child personality (peer-reported)
& 38/30
& 65
& $\sim$21.6h
& 1 $^\phi$
& English \\ \hline

\textbf{UDIVA} (ours), 2020
& Social interaction analysis
& $\frac{1}{5}$ Non-acted \& $\frac{4}{5}$Non-acted$^*$ 
& Audiovisual, heart rate 
& Personality scores (self- \& peer- reported), sociodemographics, mood, fatigue, relationship type
& 66/81
& 188$\times$5 (tasks)
& 90.5h
& 6 + 2E
& Spanish, Catalan, English \\ \hline

\multicolumn{9}{l}{$^{\dag}$ Here we consider the Green Persuasive and the EmoTABOO \cite{zara2007collection} databases together.}\\
\end{tabular}%
}
\label{tab:databases}
\end{table*}

This section reviews related work on dyadic scenarios along three axes: social signals and behaviors in context, personality recognition, and human interaction datasets. 

\textbf{Social signals and behaviors in context.} Dyadic interactions are a rich source of overt behavioral cues. They can reveal our personal attributes and cognitive/affective inner states dependent upon the context in which they are situated. Context can take many forms, and in the case of recognition of an individual state or attribute, the interaction partner’s attributes and behaviors can be considered part of the target person’s context. From a computational perspective, spatiotemporal and multiview information can be referred to as context as well. For the measurement of interpersonal constructs (e.g., synchrony~\cite{delaherche2012interpersonal}, rapport~\cite{zhao2016socially}), individual social behaviors (e.g., engagement~\cite{dermouche2019engagement}) and impressions (e.g., dominance~\cite{zhang2018facial}, empathy~\cite{park2020looking}), the joint modeling of both interlocutors and/or other sources of context has been frequently considered. However, for the task of recognizing individual attributes such as emotion and personality, context has often been misrepresented, despite recurrent claims on its importance~\cite{barrett2011context, wright2014current, vinciarelli2015open, moore2017we}. 

Recent years have seen a small surge in interlocutor-aware approaches for utterance- or turn-based emotion recognition in conversation~\cite{poria2019emotion} and sentiment analysis. Early works were based on handcrafted nonverbal, spatiotemporal dyadic features~\cite{lee2009modeling, metallinou2013tracking}. Nowadays, most approaches rely on deep learning, using conversation transcripts as input with contextualized word or speaker embeddings~\cite{lian2020context} and considering past and/or future parts of the conversation as additional context. Temporal modeling of those feature representations has been widely performed via recurrent approaches~\cite{majumder2019dialoguernn}, and more recently with BERT/Transformer-like architectures ~\cite{zhong2019knowledge, li2020hierarchical}. Some works have further proposed to enrich models with additional modalities, such as raw audiovisual data to enhance the representation of interlocutors’ influences and dynamics~\cite{zadeh2018multi,Hazarika:2018}, or speech cues in addition to the personality of the target speaker~\cite{li2019attention}. Context-aware personality recognition has followed a similar trend as for emotion, but the literature is even scarcer. We discuss it next.

\textbf{Automatic personality recognition.} Personality is widely defined as the manifestation of individual differences in patterns of thought, feeling, and behavior, that remain relatively stable during time~\cite{Soto:John:2017}. In the personality computing field~\cite{Vinciarelli:TAC2013}, it is usually characterized by the basic Big Five traits~\cite{McCrae:1992} (\textit{Openness to Experience}, \textit{Conscientiousness}, \textit{Extraversion}, \textit{Agreeableness}, and \textit{Neuroticism}), often referred to as OCEAN, based on self-reported assessments. Most works focus on personality recognition from the individual point of view, even in a dyadic or small group conversational context~\cite{aran2013cross}, using only features from the target person. Preliminary studies tended to use handcrafted features from gestures and speech~\cite{Nguyen:ICMI:2013}, while more recent works rely on deep learning approaches from raw data~\cite{mehta2019recent}.

To our knowledge, few methods propose interlocutor- or context-aware methods for personality recognition. The work of~\cite{su2016exploiting} was one of the first, leveraging turn-taking temporal evolution from transcript features but focusing on apparent personality recognition (i.e., personality reported by external observers~\cite{Jacques:TAC:2019}). With respect to self-reported personality inference in small group interactions, ~\cite{fang2016personality} regressed individual and dyadic features of personality and social impressions utilizing handcrafted descriptors of prosody, speech and visual activity. Later, ~\cite{lin2018using} proposed an interlocutor-modulated recurrent attention model with turn-based acoustic features. Finally, ~\cite{zhang2020multiparty} predicted personality and performance labels by correlation analysis of co-occurrent key action events, which were extracted from head and hand pose, gaze and motion intensity features. Regarding context, just one previous approach added person metadata (e.g., gender, age, ethnicity, and perceived attractiveness) to audiovisual data ~\cite{principi2019effect}. However, their goal was to better approximate the crowd biases for apparent personality recognition in one-person videos. Contrary to previous works, we use different sources of context, including both interlocutors, scene, and task information to infer personality, using for the first time a video-based transformer adapted to include audio and further context as metadata.

\textbf{Human interaction datasets.} Research on human behavior and communication understanding has fostered the creation of a plethora of human interaction datasets~\cite{kipp2009multimodal, poria2017review, dudzik2019context, stergiou2019analyzing}. Here, we focus on publicly available datasets containing at least audiovisual data, which enable the fusion of multiple modalities and the creation of more complete representations. In the literature, we can find examples of rich, non-acted datasets focused on computer-mediated dyadic settings~\cite{cafaro2017noxi, kossaifi2019sewa}, face-to-face triadic~\cite{Joo_2019_CVPR, celiktutan2017multimodal}, or small group interactions~\cite{alameda2015salsa}. A number of TV-based datasets with acted interactions also exist~\cite{poria2019meld}. However, in such cases, the interlocutors’ internal states are artificially built.

One of the advantages of face-to-face settings is that the full overt behavioral spectrum can be observed and modeled. Existing publicly available face-to-face dyadic interaction datasets are summarized in Table~\ref{tab:databases}\footnote{The complete table is included in the supplementary material.}. 

As it can be seen, most of them are limited in the number of participants, recordings, views, context annotations, language, or purpose. The UDIVA dataset has been designed with a multipurpose objective and aims at filling this gap.

\vspace{-0.2cm}

\section{UDIVA dataset}
\label{sec:dataset}
This section introduces the UDIVA dataset (Understanding Dyadic Interactions from Video and Audio signals), consisting of time-synchronized multimodal, multiview videos of non-scripted face-to-face dyadic interactions based on free and structured tasks performed in a lab setup\footnote{Additional details regarding design, participants recruitment, technical setup, and descriptive statistics will be provided in a follow-up paper.}.

\begin{figure*}[t!]
    \centering    
    \begin{subfigure}[]
        {\includegraphics[height=3.38cm]{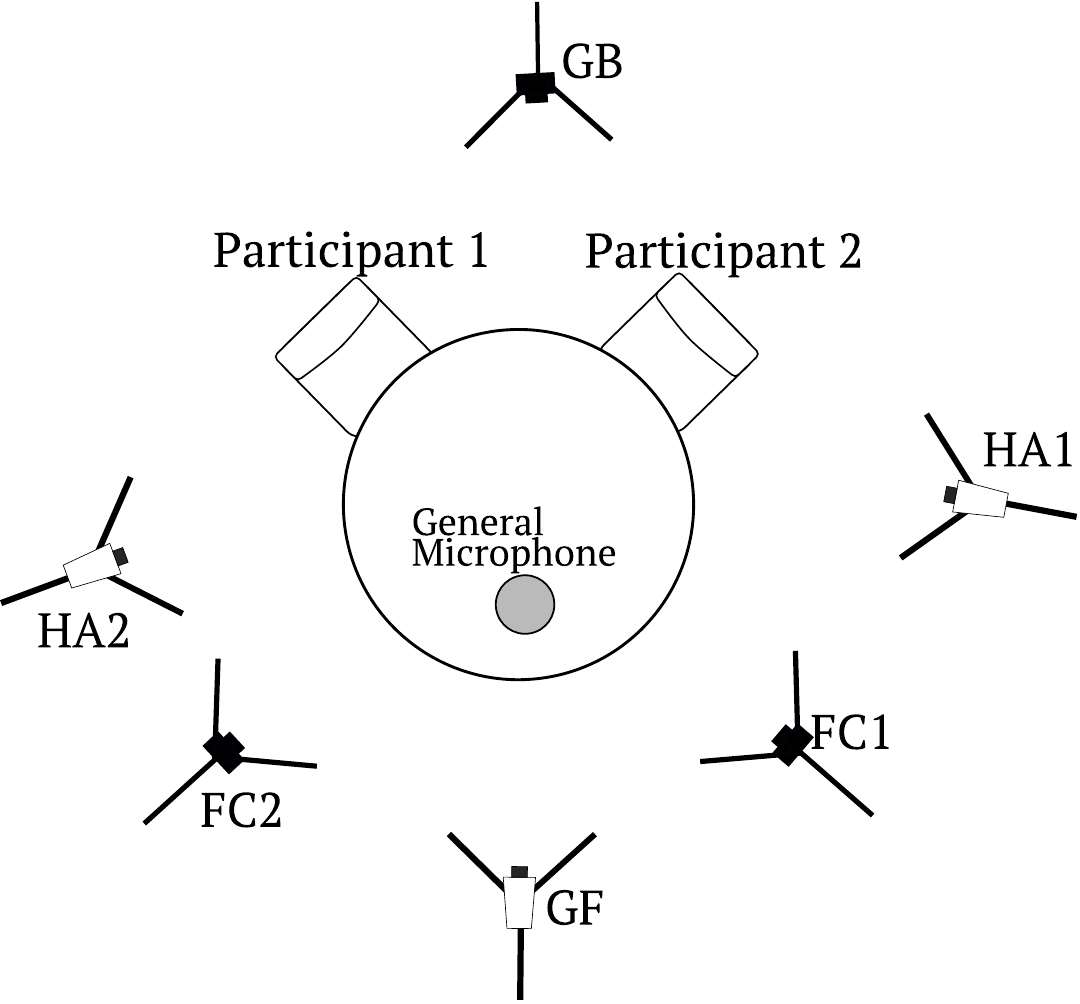} 
        \label{fig:settingarchitecture}}
    \end{subfigure}    \hfill    \begin{subfigure}[]
        {\includegraphics[height=3.38cm]{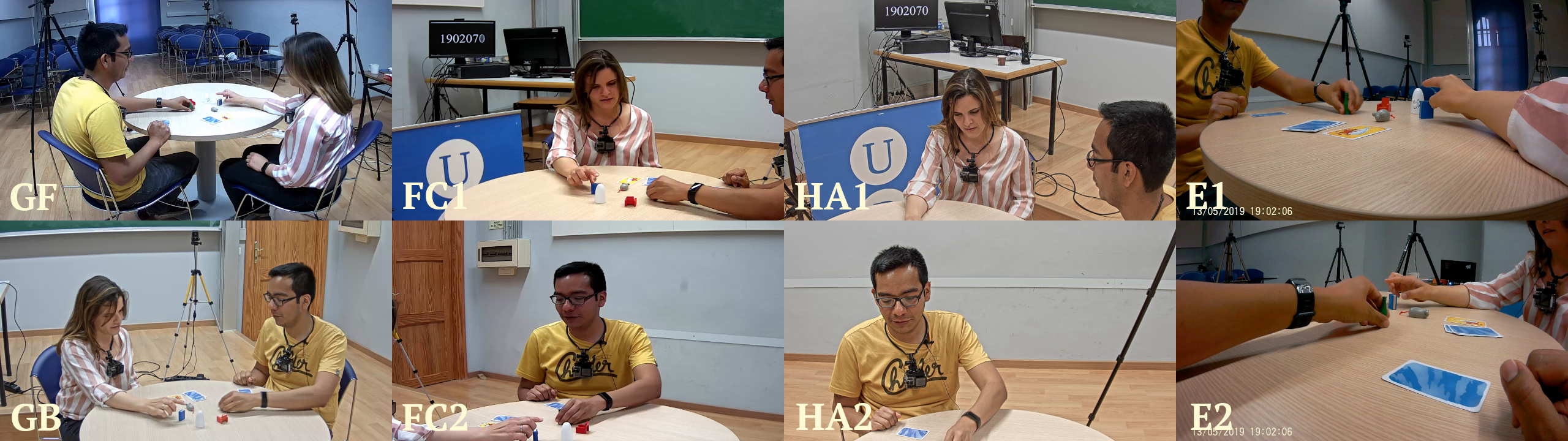}
        \label{fig:camviews}}
    \end{subfigure}
    \caption{Recording environment. We used six tripod-mounted cameras, namely \textbf{GB}: General Rear camera, \textbf{GF}: General Frontal camera, \textbf{HA}: individual High Angle cameras and \textbf{FC}: individual Frontal Cameras, and two ego cameras \textbf{E} (one per participant, placed around their neck). a) Position of cameras, general microphone and participants. b) Example of the time-synchronized 8 views.\vspace{-0.1cm}}
    \label{fig:viewsandarchitecture}
\end{figure*}

\subsection{Motivation} 
\vspace{-0.1cm}

UDIVA wants to move beyond automatic individual behavior detection and focus on the development of automatic approaches to study and understand the mechanisms of influence, perception and adaptation to verbal and nonverbal social signals in dyadic interactions, taking into account individual and dyad characteristics as well as other contextual factors. One of our research questions centers on the feasibility of developing systems able to unravel the personality 
and internal processes of an individual by the social signals they convey, and to understand how interaction partners perceive and react to those cues directed to them. By publicly releasing the dataset to the research community, we encourage data sharing and collaboration among different disciplines, reuse, and repurposing of new research questions.

\begin{figure*}[t!]
\centering    
\includegraphics[width=\textwidth]{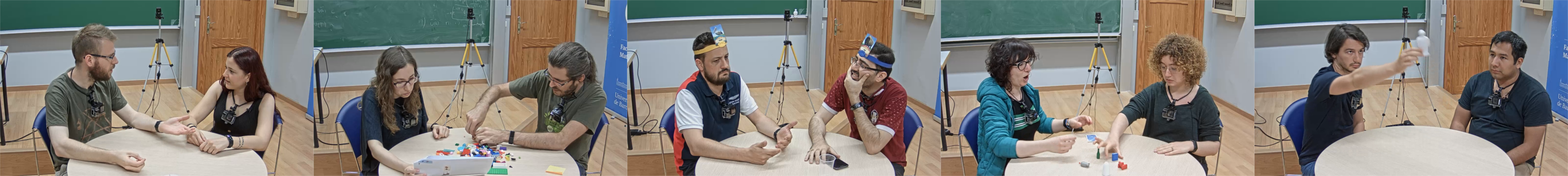}
\caption{Examples of the 5 tasks included in the UDIVA dataset from 5 sessions. From left to right: \textit{Talk}, \textit{Lego}, \textit{Animals}, \textit{Ghost}, \textit{Gaze}.}
\label{fig:tasks}\vspace{-0.3cm}
\end{figure*}

\vspace{-0.1cm}\subsection{Main statistics}
\vspace{-0.1cm}
The dataset is composed of 90.5h of recordings of dyadic interactions between 147 voluntary participants\footnote{Participants gave consent to be recorded and to share their collected data for research purposes, in compliance with GDPR \url{https://ec.europa.eu/info/law/law-topic/data-protection_en}.} (55.1\% male) from 4 to 84 years old (mean=31.29), coming from 22 countries (68\% from Spain). The majority of participants were students (38.8\%), and identified themselves as white (84.4\%). Participants were distributed into 188 dyadic sessions, with a participation average of 2.5 sessions/participant (max. 5 sessions). To create the dyads, three variables were taken into account: 1) gender (\textit{Female}, \textit{Male}); 2) age group (\textit{Child}: 4-18, \textit{Young}: 19-35, \textit{Adult}: 36-50, and \textit{Senior}: 51-84); and 3) relationship among interlocutors (\textit{Known}, \textit{Unknown}). Participants were matched according to their availability and language while trying to enforce a close-to-uniform distribution among all possible combinations between variables (60 combinations). A minimum age of 4 years and the ability to maintain a conversation in English, Spanish or Catalan were the only inclusion criteria. In the end, the most common interaction group is \textit{Male}-\textit{Male}/\textit{Young}-\textit{Young}/\textit{Unknown} (15\%), with 43\% of the interactions happening among known people. Spanish is the majority language of interaction (71.8\%), followed by Catalan (19.7\%). Half of the sessions include both interlocutors with Spain as country of origin. 

\subsection{Questionnaire-based assessments} \label{sec:questionnaires}
Prior to their first session, each participant filled a sociodemographic questionnaire, including: age, gender, ethnicity, occupation, maximum level of education, and country of origin. To assess personality and/or temperament, 
age-dependent standardized questionnaires were administered. In particular, parents of children up to 8 years old completed the Children Behavior Questionnaire (CBQ)~\cite{Rothbart:etal:2001,Osa:2013}, participants from 9 to 15 years old completed the Early Adolescent Temperament Questionnaire (EATQ-R)~\cite{Ellis:Rothbart:2001}, while participants aged 16 and older completed both the Big Five Inventory (BFI-2)~\cite{Soto:John:2017} and the Honesty-Humility axis of the HEXACO personality inventory~\cite{Ashton:2009}.

All participants (or their parents) completed pre- and post-session mood (\cite{Gallardo-Pujol:2012}) and fatigue (ad hoc 1-to-10 rating scale) assessments. The mood assessment contained items drawn from the Post Experimental Questionnaire of Primary Needs (PEQPN~\cite{williams2002investigations}). After each session, participants aged 9 and above completed again the previous temperament/personality and mood questionnaires, this time rating the individual they interacted with, to provide their perceived impression. Finally, participants reported the relationship they had with their interaction partner, if any.

\begin{figure*}[t!]
\centering    

\includegraphics[width=\textwidth]{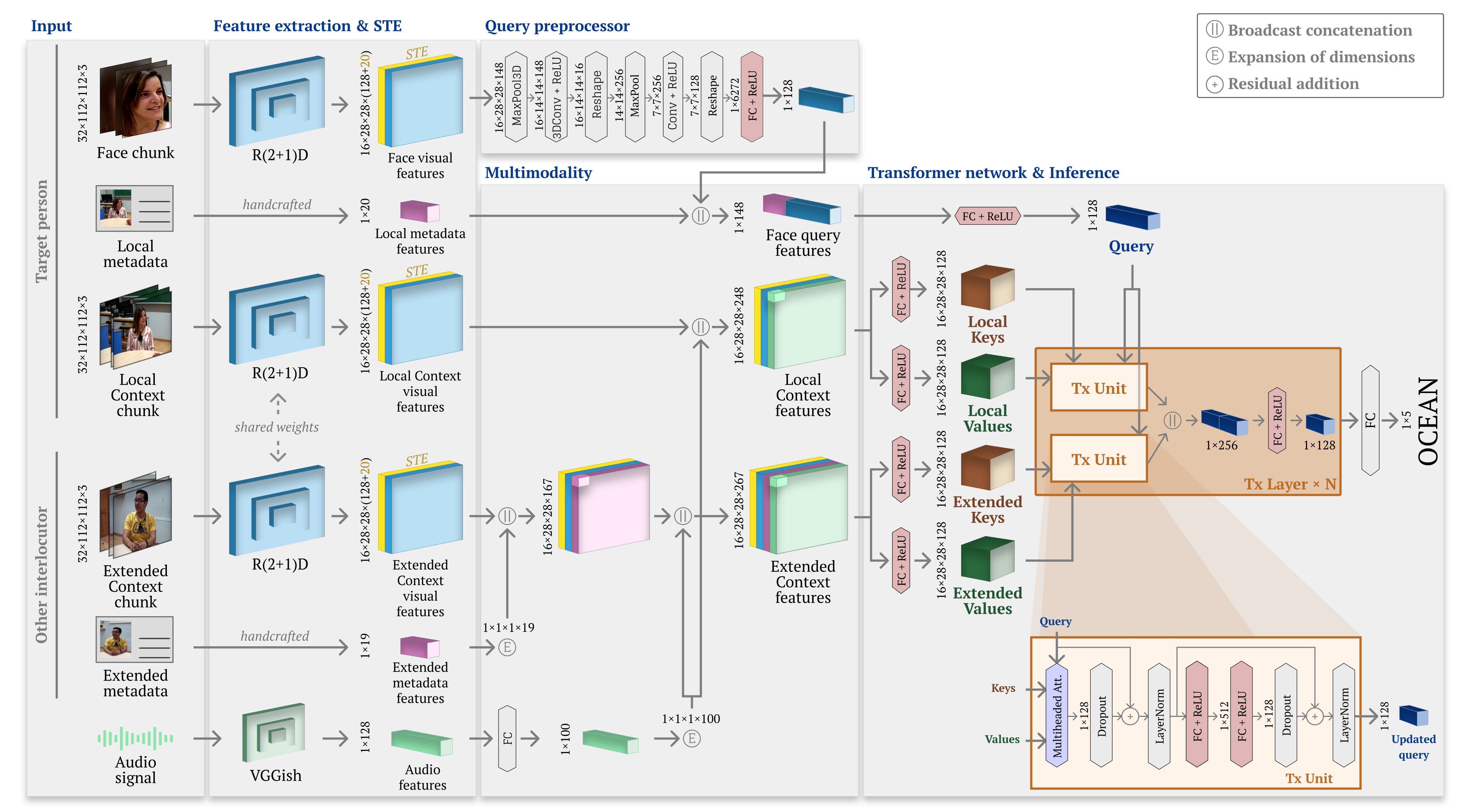}
\caption{Proposed method to infer self-reported personality (OCEAN) traits from multimodal synchronized signals and context. Input consists of visual (face, local context, and extended context chunks), audio (raw chunks), and metadata (both interlocutors' characteristics, and session and dyadic features). Feature extraction is performed by a R(2+1)D network for the visual chunks and VGGish for audio. The visual features from the R(2+1)D's 3rd residual block are concatenated to spatiotemporal encodings (STE). The VGGish's audio features and handcrafted metadata features are incorporated to visual context/query features and the result transformed to the set of Query, Keys, and Values as input to the Transformer network (Tx). The latter consists of $N$ Tx layers, each equipped with Local and Extended Context Transformer Tx units. Such units implement multiheaded attention and provide their updated queries, which are combined and fed to the next Tx Layer. Finally, the output of the $N$-th Tx layer is fed to a fully-connected (FC) layer to regress per-chunk OCEAN scores. }\vspace{-0.3cm}
\label{fig:pipeline}
\end{figure*}

\subsection{Structure of a dyadic session} \label{sec:tasks}

Participants were asked to sit at 90$^{\circ}$ to one another around a table (see Fig.~\ref{fig:settingarchitecture}), to be close enough to perform the administered tasks while facilitating data acquisition. A session consisted of 5 tasks (illustrated in Fig.~\ref{fig:tasks}) eliciting distinct behaviors and cognitive workload: 

\textbf{Talk.} Participants were instructed to talk about any subject 
during approx. 5 minutes. This task allows analysis of common conversation constructs, such as turn-taking, synchrony, empathy and quality of interaction, among others. 
    
\textbf{``Animals'' game.} Participants asked 10 \textit{yes}/\textit{no} questions each to guess the animal they had on their forehead. Animals were classified into 3 difficulty levels. This game reveals cognitive processes (e.g., thinking, gaze events). 
    
\textbf{Lego building.} Participants built a Lego together following the instructions leaflet, ranging between 4 difficulty levels. This task fosters collaboration, cooperation, joint attention, and leader-follower behaviors, among others. 
    
\textbf{``Ghost blitz'' card game.} Participants had to select, from a set of 5 figures, the one whose color and shape was not shown in a selected card. They played 1 card per turn, competing with each other to be the first at selecting the correct figure. This task fosters competitive behavior, and allows cognitive processing speed analysis, among others. 
    
\textbf{Gaze events.} Participants followed directions to look \textit{at other's face}, \textit{at static/moving object}, or \textit{elsewhere}, while moving head and eyes. This task serves as ground truth for gaze gestures and face modeling with varied head poses.

These tasks were selected along with psychologists due to the variety of individual and dyadic behaviors they elicit. In particular, Lego structures have been widely used in observational settings to assess aspects as communication~\cite{abel2017cognitive}, social skills~\cite{lindsay2017scoping} or teamwork abilities and performance~\cite{fusaroli2016heart}. \textit{Ghost} and \textit{Animals} are examples of board games, proven to be valid assessments of interpersonal skills~\cite{hartup1993conflict, underwood2004observational}. All these aspects are, in turn, indicators of personality traits like \textit{Extraversion}, \textit{Agreeableness} or \textit{Conscientiousness}~\cite{ajzen1987attitudes}. Cognitive methods, such as the tasks herein used, are routinely used in personality research~\cite{aschwanden2020cognitive}. 

The tasks were explained by a lab proctor prior to each task, who left the recording room while it was taking place. Only for \textit{Gaze} the proctor gave the instructions while participants performed them. \textit{Talk} was always administered first as a warm-up, while \textit{Gaze} was always last. The rest were administered randomly. The difficulty of \textit{Lego} and \textit{Animals} for each session was selected such that no participants repeated the same Lego or animal twice, while forcing a uniform distribution on the number of times each item was used for the total of sessions. To assess their difficulty level, we conducted an anonymous survey among 19 co-researchers. 

\vspace{-0.1cm}\subsection{Technical setup}

The setup consisted of 6 HD tripod-mounted cameras ($1280\times720$px, 25fps), 1 lapel microphone per participant and an omnidirectional microphone on the table, as depicted in Fig.~\ref{fig:settingarchitecture}. Each participant also wore an egocentric camera ($1920\times1080$px, 30fps) around their neck and a heart rate monitor on their wrist. All the capturing devices are time-synchronized and the tripod-mounted cameras calibrated. See Fig.~\ref{fig:camviews} for an example of the camera views.

\section{Personality traits inference}
\label{sec:method}
\newenvironment{tight_enumerate}{
\begin{enumerate}
  \setlength{\itemsep}{1pt}
  \setlength{\parskip}{1pt}
}{\end{enumerate}}

This section provides a first insight into the UDIVA dataset by evaluating it in a personality traits inference task. We present a transformer-based context-aware model to regress self-reported personality traits of a target person during a dyadic interaction. Then, we assess its performance and the effect of adding several sources of context. Method, evaluation protocol and results are described next\footnote{Additional details are provided in the supplementary material.}.

\begin{subsection}{Method description}
\vspace{-0.1cm}

The attention mechanism of our transformer-based method relates an initial query, in this case the target person's face, to the nonverbal behavior of both interlocutors, the overall scene, and further contextual metadata, and updates it with relevant context. The process is repeated with the updated query in consecutive layers to eventually infer the personality (OCEAN) traits. The proposed method consists of several stages, detailed below. All components and the information flow among them are illustrated in Fig.~\ref{fig:pipeline}.

\textbf{Audiovisual input.} Let $\mathbf{X}_L, \mathbf{X}_E \in [0,255]^{\mathcal{L} \times H \times W \times 3}$ be the pair of time-synchronized full-length videos containing the target person (\textit{local context}) and the other interlocutor (\textit{extended context}), respectively. We divide them into 32-frame non-overlapping chunks and resize each chunk's spatial resolution to $112 \times 112$ to obtain, respectively, $\mathbf{B}_L, \mathbf{B}_E \in [0,255]^{32 \times 112 \times 112 \times 3}$. The 32 frames of the chunks are sampled with a stride of 2, such that a chunk corresponds to approx. 2.5 seconds of the original videos. Also, we detect the target person's face regions in $\mathbf{X}_L$, crop, and re-scale them to form the face chunk $\mathbf{B}_F \in [0,255]^{32 \times 112 \times 112 \times 3}$. As face detector, we use a MobileNet-SSD~\cite{howard2017mobilenets} model pretrained on Widerface~\cite{yang2016wider}. Apart from the visual data, we define an audio chunk $\mathbf{b}_A \in \mathbb{R}^{132\,300}$ consisting of the raw audio frames acquired at 44.1 KHz from the general microphone (or one of the lapels if the general one was not available for that session), and time-synchronized to its respective video chunk.

\textbf{Metadata input}. Different sources of context are captured in the form of input metadata, described in Table~\ref{tab:metadata}. We consider 2 types of metadata: (1) \textit{local metadata}, containing \textit{individual} context from the target person and \textit{session} information; and (2) \textit{extended metadata}, with \textit{individual} context from the other interlocutor and \textit{dyadic} features. 

\textbf{Feature extraction.} First, we normalize the pixel values of $\{\mathbf{B}_F, \mathbf{B}_L, \mathbf{B}_E\}$ in the range $[0,1]$, subtracting and dividing them by the mean and standard deviation of the IG-65M dataset~\cite{ghadiyaram2019large}. Then, we feed them to a R(2+1)D network~\cite{tran2018closer} backbone, pretrained on that same dataset, and save the rich spatiotemporal features produced by the R(2+1)D's 3rd convolutional residual stack: $\mathbf{Z}'_F =g_F(\mathbf{B}_F; \theta_F),\; \mathbf{Z}'_L = g_L(\mathbf{B}_L; \theta_C),\; \mathbf{Z}'_E = g_E(\mathbf{B}_E; \theta_C)$, where $\theta_F$ are the weights of the face network $g_F(\cdot)$, and $\theta_C$ are shared weights of $g_L(\cdot)$ and $g_E(\cdot)$ networks. $\mathbf{Z}'_F, \mathbf{Z}'_L, \mathbf{Z}'_E \in \mathbb{R}^{16 \times 28 \times 28 \times 128}$ denote the \textit{face}, \textit{local context}, and \textit{extended context visual features}, respectively. For the audio feature extraction, we use the VGGish~\cite{45611} backbone. This VGG-like model, developed specifically for the audio modality and with pre-trained weights $\theta_A$ learned on a preliminary version of the YouTube-8M~\cite{DBLP:journals/corr/Abu-El-HaijaKLN16}, provides a feature vector $\mathbf{a} \in \mathbb{R}^{128}$ encoding information contained in the $\mathbf{b}_A$ chunk: $\mathbf{a}=g_A(\mathbf{b}_A; \theta_A)$. Finally, input metadata is normalized according to Table~\ref{tab:metadata}, and encoded in $\mathbf{m}_L \in \mathbb{R}^{20}$ and $\mathbf{m}_\mathrm{E} \in \mathbb{R}^{19}$ for \textit{local} and \textit{extended metadata features}, respectively.

\begin{table}[t!]
\caption{Description of the different sources of context included as metadata in the proposed personality inference model.}
\centering
\resizebox{\linewidth}{!}{
\begin{tabular}{|M{0.5cm}|M{1.5cm}|M{2cm}|M{2.5cm}|M{2.7cm}|M{1.5cm}|} \hline
    \multicolumn{3}{|c|}{\textbf{Context type}} & \textbf{Source} & \shortstack{\\ \textbf{Value range} \\ \textbf{normalization}} & \shortstack{\textbf{Output} \\ \textbf{size}} \\ \hline
    \multirow{11}{*}{\rotatebox[origin=c]{90}{\textbf{Individual}}}
    & \multirow{5}{*}{\shortstack{Stable \\ (across \\ sessions)}} 
    & Age & Self-reported &  $[17,75]\to[0,1] $ & 1D  \\ \cline{3-6} 
     &  & Gender & Self-reported & $\{\text{F},\text{M}\}\to\{0,1\}$  &  1D \\ \cline{3-6} 
     &  & \shortstack{Cultural\\ background} & \shortstack{Self-reported\\ (country of origin)} & \shortstack{\\ Recategorization \\ based on cultural \\ differences~\cite{mensah2013global}}  & \shortstack{6D \\ (one-hot \\ encoding)} \\ 
    \cline{2-6} 
     & \multirow{6}{*}{\shortstack{Transient\\ (per \\ session)}} & \shortstack{\\Session\\  index} & \shortstack{\\Session \\ info.} & $[1,5] \to [0,1]$ & 1D  \\ \cline{3-6} 
     &  & \shortstack{Pre-session \\ mood} & \shortstack{\\Self-reported~\cite{Gallardo-Pujol:2012}\\ (8 categories$^*$,\\ Likert scale)} & \shortstack{$[1,5]\to[0,1]$ \\ (for each category)}& 8D \\ \cline{3-6} 
     &  & \shortstack{Pre-session\\ fatigue} & \shortstack{\\Self-reported\\ (Rating scale)} & $[0^{\dag},10]\to[0,1]$ & 1D  \\ \hline

    \multirow{2}{*}{\rotatebox[origin=c]{90}{\textbf{Session}}} & 
    \multicolumn{2}{c|}{\begin{tabular}{M{3cm}} Order of the task \\ within the session \end{tabular}}
    & \shortstack{Session \\ info.} & $[1,4] \to [0,1]$  & 1D  \\ \cline{2-6} 
     & \multicolumn{2}{c|}{\begin{tabular}{M{3cm}}Task difficulty$^\dagger$\end{tabular}} & \shortstack{\\External \\ survey} & $[0,3]\to[0,1]$ & 1D \\ \hline
     
    \rotatebox[origin=c]{90}{\textbf{~~Dyadic~}} & 
    \multicolumn{2}{c|}{\begin{tabular}{M{3cm}}\\ Interlocutors' \\ relationship\end{tabular}} & \shortstack{Self-reported}  & $\{\text{N},\text{Y}\}\to\{0,1\}$  & 1D \\ \hline
    
    \multicolumn{6}{l}{$^*$Categories: \textit{good}, \textit{bad}, \textit{happy}, \textit{sad}, \textit{friendly}, \textit{unfriendly}, \textit{tense}, and \textit{relaxed}.} 
    \\
    \multicolumn{6}{l}{$^{\dag}$ Sessions with fatigue data missing were assigned a value of 0.}
    \\
    \multicolumn{6}{l}{$^\ddagger$ Tasks with no difficulty level associated were assigned a value of 0.}
\end{tabular}
}\vspace{-0.5cm}
\label{tab:metadata}
\end{table}

\textbf{Spatiotemporal encodings (STE).} Following other transformer-like architectures, we need to add positional encodings to our audiovisual feature embeddings $\mathbf{Z}'$, which can be either learned or fixed. We opt to learn them end-to-end. Being 16 the size of the temporal dimension of the different $\mathbf{Z}'$, we create a vector of zero-centered time indices $\mathbf{t} = \left<-\frac{16}{2}, -\frac{16}{2}+1, \ldots, \frac{16}{2}-1\right>$. The \textit{temporal encodings} are computed by a two-layered network: $\mathbf{P}'_T = \mathrm{ReLU}\left(\Theta_{T_1}^{\top} \mathrm{ReLU}\left(\Theta_{T_2}^{\top} \mathbf{t}\right)\right)$, where $\Theta_{T_1} \in \mathbb{R}^{1 \times 20}$ and $\Theta_{T_2} \in \mathbb{R}^{20 \times 10}$ are learned weights. The \textit{spatial encodings} $\mathbf{P}'_S$ are computed by a similar encoding network. Given that $28 \times 28$ is the spatial resolution of the features, we feed to the spatial encoding network a tensor of spatially zero-centered position indices $\mathbf{S} \in \mathbb{R}^{28 \times 28 \times 2}$, where $\mathbf{S}_{i, j} = \left<i-\frac{28}{2}, j-\frac{28}{2}\right>, \forall i,j \in [0, 28)$ and weights $\Theta_{S_1} \in \mathbb{R}^{2 \times 20}$ and $\Theta_{S_2} \in \mathbb{R}^{20 \times 10}$. Then, $\mathbf{P}'_T$ and $\mathbf{P}'_S$ are reshaped to $\mathbf{P}_T \in \mathbb{R}^{16 \times 1 \times 1 \times 10}$ and $\mathbf{P}_S \in \mathbb{R}^{1 \times 28 \times 28 \times 10}$ and concatenated together by broadcasting singleton dimensions, i.e. $\mathbf{P} = \mathbf{P}_S \mathbin\Vert \mathbf{P}_T$. $\mathbf{P} \in \mathbb{R}^{16 \times 28 \times 28 \times 20}$ is concatenated to each of the feature embeddings $\mathbf{Z}'$: $\mathbf{Z}_F = \mathbf{Z}'_F \mathbin\Vert \mathbf{P},\; \mathbf{Z}_L = \mathbf{Z}'_L \mathbin\Vert \mathbf{P},\; \mathbf{Z}_E = \mathbf{Z}'_E \mathbin\Vert \mathbf{P},\;$ resulting in $\mathbf{Z}_F, \mathbf{Z}_L, \mathbf{Z}_E \in \mathbb{R}^{16 \times 28 \times 28 \times 148}$. To these features with spatiotemporal encodings, $\mathbf{Z}$, we will later concatenate metadata and audio to obtain the \textit{face query}, \textit{local context}, and \textit{extended context features}.  

\textbf{Query Preprocessor (QP).} This small module transforms $\mathbf{Z}_F$ to a vector-form: $\mathbf{f} = \mathrm{QP}(\mathbf{Z}_F),\; \mathbf{f} \in \mathbb{R}^{128}$. The QP consists of a 3D max pooling layer of size $(1,2,2)$ and stride $(1,2,2)$, a 3D conv layer of size $(1, 1, 1)$ and $16$ filters, a ReLU activation function layer, a permutation of dimensions and reshaping so that the temporal dimensions and the channels are merged into the same dimension, a 2D max pooling of size $(2,2)$, a 2D conv layer of size $(1,1)$, a ReLU activation layer, a flattening, and a fully-connected (FC) layer of size $128$, another ReLU, and a dropout layer. 

\textbf{Multimodality: fusing visuals with audio and metadata.} Both \textit{local} and \textit{extended visual context features} along with encodings, $\mathbf{Z}_L$ and $\mathbf{Z}_E$, are augmented with audio features. The original 128-dimensional global \textit{audio features} $\mathbf{a}$ are linearly projected to a more compact 100-dimensional representation and reshaped to $\mathbf{A} \in \mathbb{R}^{1 \times 1 \times 1 \times 100}$. Then, the \textit{local context features} are simply $\mathbf{W}_L = \mathbf{Z}_L \mathbin\Vert \mathbf{A}$. The \textit{extended context features} are augmented with the updated audio features and the \textit{extended metadata} from the interlocutor, reshaping $\mathbf{m}_E \in \mathbb{R}^{19}$ to $\mathbf{M}_E \in \mathbb{R}^{1 \times 1 \times 1 \times 19}$ and applying broadcast concatenation, that is $\mathbf{W}_E = \mathbf{Z}_E \mathbin\Vert \mathbf{A} \mathbin\Vert \mathbf{M}_E$. Finally, the \textit{face query features} $\mathbf{w}_Q \in \mathbb{R}^{148}$ are built from the combination of the QP output along with the target person's \textit{local metadata}: $\mathbf{w}_Q = \mathbf{f} \mathbin\Vert \mathbf{m}_L$.

\textbf{Keys, Values, and Query.} To obtain the final input to the transformer layers, we first need to transform \textit{local} and \textit{extended context features} into two different 128-dimensional embeddings (Keys and Values), and also the \textit{face query features} into the query embedding of the same size. The \textit{Local keys} and \textit{Local values} are then $\mathbf{K}_L = \mathrm{ReLU}( \mathbf{\Theta}_{K_L}^{\top}\mathbf{W}_L)$ and $\mathbf{V}_L = \mathrm{ReLU}(\mathbf{\Theta}_{V_L}^{\top}\mathbf{W}_L)$ where $\mathbf{\Theta}_{K_L}, \mathbf{\Theta}_{V_L}\in \mathbb{R}^{248 \times 128}$, whereas the \textit{Extended keys} and \textit{Extended values} are $\mathbf{K}_E = \mathrm{ReLU}(\mathbf{\Theta}_{K_E}^{\top}\mathbf{W}_E)$ and $\mathbf{V}_E = \mathrm{ReLU}(\mathbf{\Theta}_{V_E}^{\top}\mathbf{W}_E)$, where $\mathbf{\Theta}_{K_E}, \mathbf{\Theta}_{V_E}\in \mathbb{R}^{267 \times 128}$. The input \textit{Query} representation $\mathbf{q}_0 \in \mathbb{R}^{128}$ is computed as $\mathbf{q}_0 = \mathrm{ReLU}(\mathbf{\Theta}_{Q_0}^{\top}\mathbf{w}_Q$), where $\mathbf{\Theta}_{Q_0} \in \mathbb{R}^{148 \times 128}$.

\textbf{Transformer network.} Our transformer network (Tx) is composed of $N = 3$ Tx layers with 2 Tx units each, one for the local context and another one for the extended context. The units consist of a multiheaded attention layer with $H = 2$ heads each. Each head computes a separate $128/H$-dimensional linear projection of the query, the keys, and the values, and applies scaled dot product attention as in~\cite{NIPS2017_7181}. Then, it concatenates the $H$ outputs, and linearly projects them back to a new 128-dimensional query. After the multiheaded attention, the resulting query follows the rest of the pipeline in the Tx unit (as illustrated in Fig.~\ref{fig:pipeline}) to obtain the \emph{updated query}. Note that each unit in the $i$-th layer provides its own updated query, denoted as $\mathbf{q}_{L_i} \in \mathbb{R}^{128}$ and $\mathbf{q}_{E_i} \in \mathbb{R}^{128}$, $0 < i \leq N$. These are next concatenated together and fed to a FC layer to obtain the $i$-th layer's joint updated query $\mathbf{q}_i = \mathrm{ReLU}\left(\mathbf{\Theta}_{Q_i}^{\top} (\mathbf{q}_{\mathrm{L}_i} \mathbin\Vert \mathbf{q}_{\mathrm{E}_i}) \right)$, where $\mathbf{\Theta}_{Q_i} \in \mathbb{R}^{256 \times 128}$. Finally, $\mathbf{q}_i$ is fed as input to the next ($i+1$-th) layer. 

\textbf{Inference.} The per-chunk OCEAN traits are obtained by applying a FC layer to the updated query from the $N$-th (last) layer, i.e. $\mathbf{y} =\mathbf{\Theta}_{\mathrm{FC}}^{\top} \mathbf{q}_N$ where $\mathbf{\Theta}_{\mathrm{FC}} \in \mathbb{R}^{128 \times 5}$. Final per-trait, per-subject predictions are computed as the median of the chunks predictions for each participant.

\end{subsection}
\begin{subsection}{Experimental setup} \label{sec:exp_eval} 
\vspace{-0.1cm}

\newcommand{\cmark}{\ding{51}}
\begin{table}[t!]
\caption{Evaluated scenarios. Mean value baseline~(B) obtained from the mean of the per-trait ground truth labels of the training set; and the proposed method with/without Local~(L) and Extended~(E) context, Metadata~(m), and Audio~(a) information.}
\resizebox{1.0\linewidth}{!}{%
\begin{tabular}{c|c|c!{\vrule width 1pt}c|c|c|c|}
\cline{2-7}
 & \multicolumn{2}{c!{\vrule width 1pt}}{\textbf{Query}} & \multicolumn{4}{c|}{\textbf{Key} and \textbf{Value}} \\ \cline{2-7}
\multicolumn{1}{c|}{} & Face$^*$ & Metadata$^*$ & Frame$^*$ & Frame$^\ddagger$ & Metadata$^\ddagger$ & Audio \\ \hline
\multicolumn{1}{|c|}{B} & - & - & - & - & - & - \\ \hline
\multicolumn{1}{|c|}{L} & \cmark & - & \cmark & - & - & - \\ \hline
\multicolumn{1}{|c|}{Lm} & \cmark & \cmark & \cmark & - & - & - \\ \hline
\multicolumn{1}{|c|}{LE} & \cmark & - & \cmark & \cmark & - & - \\ \hline
\multicolumn{1}{|c|}{LEm} & \cmark & \cmark & \cmark & \cmark & \cmark & - \\ \hline
\multicolumn{1}{|c|}{LEa} & \cmark & - & \cmark & \cmark & - &  \cmark \\ \hline
\multicolumn{1}{|c|}{LEam} & \cmark & \cmark & \cmark & \cmark & \cmark & \cmark \\ \hline
\multicolumn{7}{l}{$^*$ target person and $^\ddagger$ interlocutor data.}
\end{tabular}%
}\vspace{-0.6cm}
\label{tab:baseline-type}
\end{table}

\begin{table*}[t!]
\caption{Obtained results on different tasks. Legend: Mean value baseline~(B) obtained from the mean of the per-trait ground truth labels of the training set; and the proposed method with/without Local (L) and/or Extended (E) context, Metadata (m), and Audio (a) information.}
\resizebox{\textwidth}{!}{%
\begin{tabular}{c|ccccc|c!{\vrule width 1pt}ccccc|c!{\vrule width 1pt}ccccc|c!{\vrule width 1pt}ccccc|c|}
\cline{2-25}
                                       & \multicolumn{6}{c!{\vrule width 1pt}}{\textbf{Animals}}                                                                                      & \multicolumn{6}{c!{\vrule width 1pt}}{\textbf{Ghost}}                                                                                        & \multicolumn{6}{c!{\vrule width 1pt}}{\textbf{Lego}}                                                                                         & \multicolumn{6}{c|}{\textbf{Talk}}                                                                                         \\ \cline{2-25} 
                                       & \multicolumn{1}{c|}{O} & \multicolumn{1}{c|}{C} & \multicolumn{1}{c|}{E} & \multicolumn{1}{c|}{A} & N     & \textbf{Avg}   & \multicolumn{1}{c|}{O} & \multicolumn{1}{c|}{C} & \multicolumn{1}{c|}{E} & \multicolumn{1}{c|}{A} & N     & \textbf{Avg}   & \multicolumn{1}{c|}{O} & \multicolumn{1}{c|}{C} & \multicolumn{1}{c|}{E} & \multicolumn{1}{c|}{A} & N     & \textbf{Avg}   & \multicolumn{1}{c|}{O} & \multicolumn{1}{c|}{C} & \multicolumn{1}{c|}{E} & \multicolumn{1}{c|}{A} & N     & \textbf{Avg}   \\ \hline
\multicolumn{1}{|c|}{B}    & 0.731                  & 0.871                  & 0.988                  & 0.672                  & 1.206 & 0.894          & 0.733                  & 0.887                  & 0.991                  & 0.674                  & 1.220 & 0.901          & 0.738                  & 0.871                  & 0.990                  & 0.676                  & 1.204 & 0.896          & \textbf{0.731}                  & 0.872                  & 0.991                  & 0.673                  & 1.211 & 0.896          \\ \cline{1-1}
\multicolumn{1}{|c|}{L}            & 0.742                  & 0.879                  & 0.955                  & 0.674                  & 1.133 & 0.877          & 0.744                  & 0.891                  & 1.010                  & 0.677                  & 1.242 & 0.913          & \textbf{0.723}                  & 0.852                  & 0.917                  & 0.676                  & 1.164 & 0.866          & 0.769                  & 0.769                  & 0.997                  & 0.664                  & 1.177 & 0.875          \\ \cline{1-1}
\multicolumn{1}{|c|}{Lm}       & \textbf{0.721}                  & 0.874                  & 0.946                  & 0.684                  & 1.154 & 0.876          & 0.759                  & 0.859                  & 1.027                  & \textbf{0.642}                  & 1.208 & 0.899          & 0.725                  & 0.798                  & 0.857                  & 0.618                  & 1.101 & 0.820          & 0.743                  & 0.798                  & 0.962                  & \textbf{0.636}                  & 1.168 & 0.861          \\ \cline{1-1}
\multicolumn{1}{|c|}{LE}           & 0.733                  & 0.832                  & 0.988                  & 0.672                  & 1.221 & 0.889          & 0.731                  & 0.905                  & 0.956                  & 0.676                  & 1.291 & 0.912          & 0.731                  & 0.885                  & 0.949                  & 0.676                  & 1.230 & 0.894          & 0.738                  & 0.793                  & 0.964                  & 0.673                  & 1.094 & 0.852          \\ \cline{1-1}
\multicolumn{1}{|c|}{LEm}      & 0.736                  & 0.834                  & 0.968                  & 0.669                  & 1.192 & 0.880          & 0.743                  & 0.944                  & 0.868                  & 0.657                  & 1.153 & 0.873          & 0.727                  & \textbf{0.763}                  & \textbf{0.826}                  & \textbf{0.611}                  & \textbf{1.037} & \textbf{0.793} & 0.825                  & \textbf{0.718}                  & 0.878                  & 0.639                  & 1.047 & 0.821          \\ \cline{1-1}
\multicolumn{1}{|c|}{LEa}      & 0.722                  & 0.827                  & 0.954                  & 0.672                  & 1.211 & 0.877          & \textbf{0.730}                  & \textbf{0.872}                  & 0.950                  & 0.672                  & 1.199 & 0.885          & 0.742                  & 0.867                  & 0.941                  & 0.672                  & 1.229 & 0.890          & 0.757                  & 0.728                  & 0.970                  & 0.664                  & 1.106 & 0.845          \\ \cline{1-1}
\multicolumn{1}{|c|}{LEam} & 0.737                 & \textbf{0.756}                  & \textbf{0.887}                  & \textbf{0.580}                  & \textbf{1.023} & \textbf{0.797} & 0.741                  & 0.893                  & \textbf{0.844}                  & 0.667                  & \textbf{1.139} & \textbf{0.857} & 0.745                  & 0.839                  & 0.953                  & 0.659                  & 1.099 & 0.859          & 0.773                  & 0.790                  & \textbf{0.869}                  & 0.670                  & \textbf{0.985} & \textbf{0.817} \\ \hline
\end{tabular}
}\vspace{-0.3cm}
\label{tab:baseline_results}
\end{table*}

This section describes the experimental setup used to assess the performance of the personality inference model. The evaluation is performed on all tasks except \textit{Gaze}, in which very few personality indicators were present due to the task design. We use frontal camera views (FC1 and FC2, see Fig.~\ref{fig:viewsandarchitecture}), in line with the proposed methodology.

As personality labels, we use the raw OCEAN scores obtained from the self-reported BFI-2 questionnaire, converted into z-scores using descriptive data from normative samples. 

\textbf{Data and splits description.} We use the subset of data composed of participants aged 16 years and above, for which Big-five personality traits are available (see Sec.~\ref{sec:questionnaires}). Subject-independent training, validation and test splits were selected following a greedy optimization procedure that aimed at having a similar distribution in each split with respect to participant and session characteristics, while ensuring that no participants appeared in different splits. 

In terms of sessions and participants, the final splits respectively contain: 116/99 for training, 18/20 for validation, and 11/15 for test. Although the validation split is larger than the test split, the latter contains a better trait balance. Since the duration of the videos is not constant throughout sessions and tasks, in order to balance the number of samples we uniformly selected around 120 chunks from each stream, based on the median number of chunks per video. The final sample of chunks contains $94\,960$ instances for training, $15\,350$ for validation and $7\,870$ for test, distributed among the 4 tasks. 

\textbf{Evaluation protocol.} We follow an incremental approach, starting from the \textit{local context}. Six different scenarios are evaluated, summarized in Table~\ref{tab:baseline-type}. We train one model per scenario and task, since each of the four tasks can elicit different social signals and behaviors (detailed in Sec.~\ref{sec:tasks}), which can be correlated to different degrees with distinct aspects of each personality trait. Results are evaluated with respect to the Mean Squared Error between the aggregated personality trait score and associated ground truth label for each individual in the test set. We also compare the results to a mean value baseline (``B''), computed as the mean of the per-trait ground truth labels of the training set. 

\end{subsection}
\begin{subsection}{Discussion of results} \label{sec:results}
\vspace{-0.1cm}
Obtained per-task results for the different scenarios are shown in Table \ref{tab:baseline_results}. We discuss some of the findings below.

\textbf{Effect of including extended (E) visual information.} The \textit{extended context} contains visual information from the other interlocutor's behaviors and surrounding scene, allowing the model to consider interpersonal influences during a chunk. By comparing ``L'' vs. ``LE'' we can observe that, on average, only \textit{Talk} benefits from the addition of the extended visual context. Trait-wise, \textit{Extraversion} improves for all tasks except for \textit{Lego}, which performs worse for all traits. This can be attributed to the fact that the interaction during this type of collaboration is more slow-paced than in other tasks. Therefore, interpersonal influences cannot be properly captured during just one chunk. In contrast, for more natural tasks such as \textit{Talk}, or fast-moving games such as \textit{Ghost}, there are many instant actions-reactions that can be observed during a single chunk, the effect of which is reflected in the improved results for those tasks. This motivates the need to extend the model to capture longer-time interpersonal dependencies, characteristic of human interactions, across a series of ordered chunks along time, to truly benefit from this extended information.

\textbf{Effect of including metadata (m) information.} The inclusion of metadata validates our intuition that personal, task, and dyadic details provide relevant information to the model to produce overall better predictions, particularly if the cases ``L''~vs.~``Lm'', ``LE''~vs.~``LEm'', and ``LEa''~vs.~``LEam'' are compared, with the largest improvement observed for \textit{Lego} (11.29\%, ``LE''~vs.~``LEm'' case). Considering the high heterogeneity and dimensionality of behaviors revealed in an interaction and their multiple meanings, these concise features appear to be beneficial to better guide the model and establish meaningful patterns in the data. Nonetheless, a systematic study would be needed to assess the effect of each feature individually. 

\textbf{Effect of including audio (a) information.} From comparing ``LE''~vs.~``LEa'' and ``LEm''~vs.~``LEam'', we observe that better results are obtained, on average, for all the tasks when audio information is considered. In line with previous literature~\cite{Vinciarelli:TAC2013}, it is clear that paralinguistic acoustic features are required to better model personality. However, the observed improvement is smaller for \textit{Lego}. One plausible reason would be the noise produced by the Lego pieces while being moved, or by the instructions book while turning its pages close to the microphones, which would interfere with the learning process. In the case of more natural routines like \textit{Talk}, the influence of audio is not as strong as we would have expected. In contrast, \textit{Animals}, another speaking-based task, obtains the best results for almost all traits when audio is considered. There is one salient difference among these two tasks that may explain this pattern. The latter elicits more individual covert thinking and cognitive processes that cannot be entirely observed from the visual modality, so most of the overt information comes from the spoken conversation. In contrast, the former elicits a larger range of visual cues which may be more relevant than acoustic features for certain traits. 

\textbf{Putting everything together.} In the last experiment (``LEam''), the model is aware of the overall contextual information. We notice that apart from \textit{Lego}, for which the audio drawbacks were already commented, all the other tasks seem to benefit from the provided knowledge, obtaining the lowest error value on average. 

\textbf{Baseline comparison.} We observe that \textit{Agreeableness}, followed by \textit{Openness}, obtain the lowest error among mean value baseline (``B'') results, indicating that ground truth labels for such traits are more concentrated. In those cases, none of the models achieve a substantial improvement over the baseline, except for \textit{Animals}, where ``LEam'' obtains an error of 0.58, the lowest overall. At the other end we find \textit{Neuroticism}, which is the trait with most spread values, but also the one for which we obtain the largest benefits with the evaluated models. In particular, the largest improvement overall (18.66\%) is given by ``LEam'' for \textit{Talk}.  \vspace{-0.1cm}
\end{subsection}

\section{Conclusion}
\label{sec:conclusion}
\vspace{-0.2cm}
This paper introduced UDIVA, the largest multiview audiovisual dataset of dyadic face-to-face non-scripted interactions. To validate part of its potential, we proposed a multimodal transformer-based method for inferring the personality of a target person in a dyadic scenario. We incrementally combined different sources of context (both interlocutors' scene, acoustic and task information) finding consistent improvements as they were added, which is consonant with human interaction research in the psychology field. 

UDIVA is currently being annotated with additional labels (e.g., transcriptions, continuous action/intention for human-object-human interaction) to 
allow for a more holistic analysis of human interaction from both individual and dyadic perspectives. From a methodological point of view, we plan to extend the proposed architecture to better capture long-term discriminative features. Nevertheless, we are releasing this data\footnote{The dataset will be available at \url{http://chalearnlap.cvc.uab.es/dataset/39/description/}.} with the purpose of advancing the research and understanding of human communication from a multidisciplinary perspective, far beyond personality analysis.

{\small \textbf{Acknowledgements.}
This work has been partially supported by the Spanish projects TIN2015-66951-C2-2-R, RTI2018-095232-B-C22, PID2019-105093GB-I00 (MINECO/FEDER, UE), CERCA Programme/Generalitat de Catalunya, and ICREA under the ICREA Academia programme. We gratefully thank Alejandro Alfonso, Rubén Ballester, Rubén Barco, Aleix Casellas, Roger Hernando, Andreu Masdeu, Carla Morral, Pablo Lázaro, Luis M. Pérez, Italo Vidal, Daniel Vilanova, all HUPBA members, and all participants for their support in the dataset collection phase. We also thank the Department of Mathematics and Informatics of Universitat de Barcelona for granting facilities access, the Data Protection and Bioethics departments for their assistance, and Noldus IT for loaning part of the recording setup.}

{\small
\bibliographystyle{unsrt}
\bibliography{ms.bib}

\begin{thebibliography}{10}

\bibitem{burgoon2007interpersonal}
Judee~K Burgoon, Lesa~A Stern, and Leesa Dillman.
\newblock {\em Interpersonal adaptation: Dyadic interaction patterns}.
\newblock Cambridge University Press, 2007.

\bibitem{braithwaite2014engaging}
Dawn~O Braithwaite and Paul Schrodt.
\newblock {\em Engaging theories in interpersonal communication: Multiple
  perspectives}.
\newblock Sage Publications, 2014.

\bibitem{vinciarelli2011bridging}
Alessandro Vinciarelli, Maja Pantic, Dirk Heylen, Catherine Pelachaud, Isabella
  Poggi, Francesca D'Errico, and Marc Schroeder.
\newblock Bridging the gap between social animal and unsocial machine: A survey
  of social signal processing.
\newblock {\em IEEE Transactions on Affective Computing}, 3(1):69--87, 2011.

\bibitem{el2006affective}
Rana El~Kaliouby, Rosalind Picard, and Simon Baron-Cohen.
\newblock Affective computing and autism.
\newblock {\em Annals of the New York Academy of Sciences}, 1093(1):228--248,
  2006.

\bibitem{ahuja2019react}
Chaitanya Ahuja, Shugao Ma, Louis-Philippe Morency, and Yaser Sheikh.
\newblock To react or not to react: End-to-end visual pose forecasting for
  personalized avatar during dyadic conversations.
\newblock In {\em 2019 International Conference on Multimodal Interaction},
  pages 74--84, 2019.

\bibitem{esposito2016modeling}
Anna Esposito and Lakhmi~C Jain.
\newblock Modeling emotions in robotic socially believable behaving systems.
\newblock In {\em Toward Robotic Socially Believable Behaving Systems-Volume
  I}, pages 9--14. Springer, 2016.

\bibitem{narayanan2013behavioral}
Shrikanth Narayanan and Panayiotis~G Georgiou.
\newblock Behavioral signal processing: Deriving human behavioral informatics
  from speech and language.
\newblock {\em Proceedings of the IEEE}, 101(5):1203--1233, 2013.

\bibitem{VINCIARELLI20091743}
Alessandro Vinciarelli, Maja Pantic, and Hervé Bourlard.
\newblock Social signal processing: Survey of an emerging domain.
\newblock {\em Image and Vision Computing}, 27(12):1743 -- 1759, 2009.

\bibitem{clark1983common}
Herbert~H Clark, Robert Schreuder, and Samuel Buttrick.
\newblock Common ground at the understanding of demonstrative reference.
\newblock {\em Journal of verbal learning and verbal behavior}, 22(2):245--258,
  1983.

\bibitem{cuperman2009big}
Ronen Cuperman and William Ickes.
\newblock Big five predictors of behavior and perceptions in initial dyadic
  interactions: Personality similarity helps extraverts and introverts, but
  hurts “disagreeables”.
\newblock {\em Journal of personality and social psychology}, 97(4):667, 2009.

\bibitem{segerstrale2018nonverbal}
Ullica Segerstrale and Peter Moln{\'a}r.
\newblock {\em Nonverbal communication: where nature meets culture}.
\newblock Routledge, 2018.

\bibitem{clore2007emotions}
Gerald~L Clore and Jeffrey~R Huntsinger.
\newblock How emotions inform judgment and regulate thought.
\newblock {\em Trends in cognitive sciences}, 11(9):393--399, 2007.

\bibitem{kenny1996models}
David~A Kenny.
\newblock Models of non-independence in dyadic research.
\newblock {\em Journal of Social and Personal Relationships}, 13(2):279--294,
  1996.

\bibitem{dudzik2019context}
Bernd Dudzik, Michel-Pierre Jansen, Franziska Burger, Frank Kaptein, Joost
  Broekens, Dirk~KJ Heylen, Hayley Hung, Mark~A Neerincx, and Khiet~P Truong.
\newblock Context in human emotion perception for automatic affect detection: A
  survey of audiovisual databases.
\newblock In {\em 2019 8th International Conference on Affective Computing and
  Intelligent Interaction (ACII)}, pages 206--212. IEEE, 2019.

\bibitem{park2020looking}
Hye~Jeong Park and Jae~Hwa Lee.
\newblock Looking into the personality traits to enhance empathy ability: A
  review of literature.
\newblock In {\em International Conference on Human-Computer Interaction},
  pages 173--180, 2020.

\bibitem{Girdhar_2019_CVPR}
Rohit Girdhar, Joao Carreira, Carl Doersch, and Andrew Zisserman.
\newblock Video action transformer network.
\newblock In {\em IEEE Conference on Computer Vision and Pattern Recognition
  (CVPR)}, June 2019.

\bibitem{rahman-etal-2020-integrating}
Wasifur Rahman, Md~Kamrul Hasan, Sangwu Lee, AmirAli Bagher~Zadeh, Chengfeng
  Mao, Louis-Philippe Morency, and Ehsan Hoque.
\newblock Integrating multimodal information in large pretrained transformers.
\newblock In {\em Proceedings of the 58th Annual Meeting of the Association for
  Computational Linguistics}, pages 2359--2369, 2020.

\bibitem{Busso2008}
Carlos Busso, Murtaza Bulut, Chi-Chun Lee, Abe Kazemzadeh, Emily Mower, Samuel
  Kim, Jeannette~N. Chang, Sungbok Lee, and Shrikanth~S. Narayanan.
\newblock {IEMOCAP}: Interactive emotional dyadic motion capture database.
\newblock {\em Language Resources and Evaluation}, 42(4):335, Nov 2008.

\bibitem{blache2017corpus}
Philippe Blache, Roxane Bertrand, Ga{\"e}lle Ferr{\'e}, Berthille Pallaud,
  Laurent Pr{\'e}vot, and St{\'e}phane Rauzy.
\newblock The corpus of interactional data: A large multimodal annotated
  resource.
\newblock In {\em Handbook of linguistic annotation}, pages 1323--1356.
  Springer, 2017.

\bibitem{douglas2007humaine}
Ellen Douglas-Cowie, Roddy Cowie, Ian Sneddon, Cate Cox, Orla Lowry, Margaret
  Mcrorie, Jean-Claude Martin, Laurence Devillers, Sarkis Abrilian, Anton
  Batliner, et~al.
\newblock The humaine database: Addressing the collection and annotation of
  naturalistic and induced emotional data.
\newblock In {\em International conference on affective computing and
  intelligent interaction}, pages 488--500. Springer, 2007.

\bibitem{douglas2011humaine}
Ellen Douglas-Cowie, Cate Cox, Jean-Claude Martin, Laurence Devillers, Roddy
  Cowie, Ian Sneddon, Margaret McRorie, Catherine Pelachaud, Christopher
  Peters, Orla Lowry, et~al.
\newblock The humaine database.
\newblock In {\em Emotion-Oriented Systems}, pages 243--284. Springer, 2011.

\bibitem{rehg2013decoding}
James Rehg, Gregory Abowd, Agata Rozga, Mario Romero, Mark Clements, Stan
  Sclaroff, Irfan Essa, O~Ousley, Yin Li, Chanho Kim, et~al.
\newblock Decoding children's social behavior.
\newblock In {\em IEEE Conference on Computer Vision and Pattern Recognition
  (CVPR)}, pages 3414--3421, 2013.

\bibitem{bilakhia2015mahnob}
Sanjay Bilakhia, Stavros Petridis, Anton Nijholt, and Maja Pantic.
\newblock The mahnob mimicry database: A database of naturalistic human
  interactions.
\newblock {\em Pattern recognition letters}, 66:52--61, 2015.

\bibitem{naim2015automated}
Iftekhar Naim, M~Iftekhar Tanveer, Daniel Gildea, and Mohammed~Ehsan Hoque.
\newblock Automated prediction and analysis of job interview performance: The
  role of what you say and how you say it.
\newblock In {\em IEEE International Conference and Workshops on Automatic Face
  and Gesture Recognition (FG)}, volume~1, pages 1--6, 2015.

\bibitem{Metallinou:2016}
Angeliki Metallinou, Zhaojun Yang, Chi-Chun Lee, Carlos Busso, Sharon Carnicke,
  and Shrikanth Narayanan.
\newblock The {USC} creativeit database of multimodal dyadic interactions: From
  speech and full body motion capture to continuous emotional annotations.
\newblock {\em Lang. Resour. Eval.}, 50(3):497--521, 2016.

\bibitem{Busso_2017}
C.~Busso, S.~Parthasarathy, A.~Burmania, M.~AbdelWahab, N.~Sadoughi, and
  E.~{Mower Provost}.
\newblock {MSP-IMPROV}: An acted corpus of dyadic interactions to study emotion
  perception.
\newblock {\em IEEE Transactions on Affective Computing}, 8(1):67--80,
  January-March 2017.

\bibitem{chen2020dyadic}
Huili Chen, Yue Zhang, Felix Weninger, Rosalind Picard, Cynthia Breazeal, and
  Hae~Won Park.
\newblock Dyadic speech-based affect recognition using dami-p2c parent-child
  multimodal interaction dataset.
\newblock In {\em Proceedings of the 2020 International Conference on
  Multimodal Interaction}, pages 97--106, 2020.

\bibitem{zara2007collection}
Aur{\'e}lie Zara, Val{\'e}rie Maffiolo, Jean~Claude Martin, and Laurence
  Devillers.
\newblock Collection and annotation of a corpus of human-human multimodal
  interactions: Emotion and others anthropomorphic characteristics.
\newblock In {\em International Conference on Affective Computing and
  Intelligent Interaction}, pages 464--475. Springer, 2007.

\bibitem{delaherche2012interpersonal}
Emilie Delaherche, Mohamed Chetouani, Ammar Mahdhaoui, Catherine Saint-Georges,
  Sylvie Viaux, and David Cohen.
\newblock Interpersonal synchrony: A survey of evaluation methods across
  disciplines.
\newblock {\em IEEE Transactions on Affective Computing}, 3(3):349--365, 2012.

\bibitem{zhao2016socially}
Ran Zhao, Tanmay Sinha, Alan~W Black, and Justine Cassell.
\newblock Socially-aware virtual agents: Automatically assessing dyadic rapport
  from temporal patterns of behavior.
\newblock In {\em International conference on intelligent virtual agents},
  pages 218--233, 2016.

\bibitem{dermouche2019engagement}
Soumia Dermouche and Catherine Pelachaud.
\newblock Engagement modeling in dyadic interaction.
\newblock In {\em 2019 International Conference on Multimodal Interaction},
  pages 440--445, 2019.

\bibitem{zhang2018facial}
Zhanpeng Zhang, Ping Luo, Chen~Change Loy, and Xiaoou Tang.
\newblock From facial expression recognition to interpersonal relation
  prediction.
\newblock {\em International Journal of Computer Vision}, 126(5):550--569,
  2018.

\bibitem{barrett2011context}
Lisa~Feldman Barrett, Batja Mesquita, and Maria Gendron.
\newblock Context in emotion perception.
\newblock {\em Current Directions in Psychological Science}, 20(5):286--290,
  2011.

\bibitem{wright2014current}
Aidan~GC Wright.
\newblock Current directions in personality science and the potential for
  advances through computing.
\newblock {\em IEEE Transactions on Affective Computing}, 5(3):292--296, 2014.

\bibitem{vinciarelli2015open}
Alessandro Vinciarelli, Anna Esposito, Elisabeth Andr{\'e}, Francesca Bonin,
  Mohamed Chetouani, Jeffrey~F Cohn, Marco Cristani, Ferdinand Fuhrmann, Elmer
  Gilmartin, Zakia Hammal, et~al.
\newblock Open challenges in modelling, analysis and synthesis of human
  behaviour in human--human and human--machine interactions.
\newblock {\em Cognitive Computation}, 7(4):397--413, 2015.

\bibitem{moore2017we}
Philip Moore.
\newblock Do we understand the relationship between affective computing,
  emotion and context-awareness?
\newblock {\em Machines}, 5(3):16, 2017.

\bibitem{poria2019emotion}
Soujanya Poria, Navonil Majumder, Rada Mihalcea, and Eduard Hovy.
\newblock Emotion recognition in conversation: Research challenges, datasets,
  and recent advances.
\newblock {\em IEEE Access}, 7:100943--100953, 2019.

\bibitem{lee2009modeling}
Chi-Chun Lee, Carlos Busso, Sungbok Lee, and Shrikanth~S Narayanan.
\newblock Modeling mutual influence of interlocutor emotion states in dyadic
  spoken interactions.
\newblock In {\em Tenth Annual Conference of the International Speech
  Communication Association}, 2009.

\bibitem{metallinou2013tracking}
Angeliki Metallinou, Athanasios Katsamanis, and Shrikanth Narayanan.
\newblock Tracking continuous emotional trends of participants during affective
  dyadic interactions using body language and speech information.
\newblock {\em Image and Vision Computing}, 31(2):137--152, 2013.

\bibitem{lian2020context}
Zheng Lian, Jianhua Tao, Bin Liu, Jian Huang, Zhanlei Yang, and Rongjun Li.
\newblock Context-dependent domain adversarial neural network for multimodal
  emotion recognition.
\newblock {\em Proc. Interspeech 2020}, pages 394--398, 2020.

\bibitem{majumder2019dialoguernn}
Navonil Majumder, Soujanya Poria, Devamanyu Hazarika, Rada Mihalcea, Alexander
  Gelbukh, and Erik Cambria.
\newblock Dialoguernn: An attentive rnn for emotion detection in conversations.
\newblock In {\em Proceedings of the AAAI Conference on Artificial
  Intelligence}, volume~33, pages 6818--6825, 2019.

\bibitem{zhong2019knowledge}
Peixiang Zhong, Di~Wang, and Chunyan Miao.
\newblock Knowledge-enriched transformer for emotion detection in textual
  conversations.
\newblock In {\em Proceedings of the Conference on Empirical Methods in Natural
  Language Processing and the 9th International Joint Conference on Natural
  Language Processing (EMNLP-IJCNLP)}, pages 165--176, 2019.

\bibitem{li2020hierarchical}
Qingbiao Li, Chunhua Wu, Zhe Wang, and Kangfeng Zheng.
\newblock Hierarchical transformer network for utterance-level emotion
  recognition.
\newblock {\em Applied Sciences}, 10(13):4447, 2020.

\bibitem{zadeh2018multi}
Amir Zadeh, Paul~Pu Liang, Soujanya Poria, Prateek Vij, Erik Cambria, and
  Louis-Philippe Morency.
\newblock Multi-attention recurrent network for human communication
  comprehension.
\newblock In {\em AAAI Conference on Artificial Intelligence}, volume 2018,
  page 5642, 2018.

\bibitem{Hazarika:2018}
Devamanyu Hazarika, Soujanya Poria, Amir Zadeh, Erik Cambria, Louis-Philippe
  Morency, and Roger Zimmermann.
\newblock Conversational memory network for emotion recognition in dyadic
  dialogue videos.
\newblock In {\em Proceedings of the Conference of the North {A}merican Chapter
  of the Association for Computational Linguistics: Human Language
  Technologies}, pages 2122--2132, 2018.

\bibitem{li2019attention}
Jeng-Lin Li and Chi-Chun Lee.
\newblock Attention learning with retrievable acoustic embedding of personality
  for emotion recognition.
\newblock In {\em 2019 8th International Conference on Affective Computing and
  Intelligent Interaction (ACII)}, pages 171--177. IEEE, 2019.

\bibitem{Soto:John:2017}
Christopher Soto and Oliver John.
\newblock The next big five inventory (bfi-2): Developing and assessing a
  hierarchical model with 15 facets to enhance bandwidth, fidelity, and
  predictive power.
\newblock {\em Journal of Personality and Social Psychology}, 113:117--143, 07
  2017.

\bibitem{Vinciarelli:TAC2013}
Alessandro Vinciarelli and Gelareh Mohammadi.
\newblock A survey of personality computing.
\newblock {\em IEEE Transaction on Affective Computing}, 5(3):273--291, 2014.

\bibitem{McCrae:1992}
Robert~R. McCrae and Oliver~P. John.
\newblock An introduction to the five-factor model and its applications.
\newblock {\em Journal of Personality}, 60(2):175--215, 1992.

\bibitem{aran2013cross}
Oya Aran and Daniel Gatica-Perez.
\newblock Cross-domain personality prediction: from video blogs to small group
  meetings.
\newblock In {\em International Conference on Multimodal Interaction}, pages
  127--130, 2013.

\bibitem{Nguyen:ICMI:2013}
Laurent~Son Nguyen, Alvaro Marcos-Ramiro, Martha Marr\'{o}n~Romera, and Daniel
  Gatica-Perez.
\newblock Multimodal analysis of body communication cues in employment
  interviews.
\newblock In {\em International Conference on Multimodal Interaction (ICMI)},
  pages 437--444, 2013.

\bibitem{mehta2019recent}
Yash Mehta, Navonil Majumder, Alexander Gelbukh, and Erik Cambria.
\newblock Recent trends in deep learning based personality detection.
\newblock {\em Artificial Intelligence Review}, pages 1--27, 2019.

\bibitem{su2016exploiting}
Ming-Hsiang Su, Chung-Hsien Wu, and Yu-Ting Zheng.
\newblock Exploiting turn-taking temporal evolution for personality trait
  perception in dyadic conversations.
\newblock {\em Transactions on Audio, Speech, and Language Processing},
  24(4):733--744, 2016.

\bibitem{Jacques:TAC:2019}
J.~C.~S. {Jacques Junior}, Y.~{Güçlütürk}, M.~{Perez}, U.~{Güçlü},
  C.~{Andujar}, X.~{Baró}, H.~J. {Escalante}, I.~{Guyon}, M.~A.~J. {Van
  Gerven}, R.~{Van Lier}, and S.~{Escalera}.
\newblock First impressions: A survey on vision-based apparent personality
  trait analysis.
\newblock {\em IEEE Transactions on Affective Computing}, pages 1--1, 2019.

\bibitem{fang2016personality}
Sheng Fang, Catherine Achard, and S{\'e}verine Dubuisson.
\newblock Personality classification and behaviour interpretation: An approach
  based on feature categories.
\newblock In {\em Proceedings of the 18th ACM International Conference on
  Multimodal Interaction}, pages 225--232, 2016.

\bibitem{lin2018using}
Yun-Shao Lin and Chi-Chun Lee.
\newblock Using interlocutor-modulated attention blstm to predict personality
  traits in small group interaction.
\newblock In {\em International Conference on Multimodal Interaction}, pages
  163--169, 2018.

\bibitem{zhang2020multiparty}
Lingyu Zhang, Indrani Bhattacharya, Mallory Morgan, Michael Foley, Christoph
  Riedl, Brooke Welles, and Richard Radke.
\newblock Multiparty visual co-occurrences for estimating personality traits in
  group meetings.
\newblock In {\em The IEEE Winter Conference on Applications of Computer
  Vision}, pages 2085--2094, 2020.

\bibitem{principi2019effect}
Ricardo Dar{\'\i}o~P{\'e}rez Principi, Cristina Palmero, Julio~C Junior, and
  Sergio Escalera.
\newblock On the effect of observed subject biases in apparent personality
  analysis from audio-visual signals.
\newblock {\em IEEE Transactions on Affective Computing}, 2019.

\bibitem{kipp2009multimodal}
Michael Kipp, Jean-Claude Martin, Patrizia Paggio, and Dirk Heylen.
\newblock {\em Multimodal corpora: from models of natural interaction to
  systems and applications}, volume 5509.
\newblock Springer, 2009.

\bibitem{poria2017review}
Soujanya Poria, Erik Cambria, Rajiv Bajpai, and Amir Hussain.
\newblock A review of affective computing: From unimodal analysis to multimodal
  fusion.
\newblock {\em Information Fusion}, 37:98--125, 2017.

\bibitem{stergiou2019analyzing}
Alexandros Stergiou and Ronald Poppe.
\newblock Analyzing human--human interactions: A survey.
\newblock {\em Computer Vision and Image Understanding}, 188:102799, 2019.

\bibitem{cafaro2017noxi}
Angelo Cafaro, Johannes Wagner, Tobias Baur, Soumia Dermouche, Mercedes
  Torres~Torres, Catherine Pelachaud, Elisabeth Andr{\'e}, and Michel Valstar.
\newblock The noxi database: multimodal recordings of mediated novice-expert
  interactions.
\newblock In {\em Proceedings of the 19th ACM International Conference on
  Multimodal Interaction}, pages 350--359, 2017.

\bibitem{kossaifi2019sewa}
Jean Kossaifi, Robert Walecki, Yannis Panagakis, Jie Shen, Maximilian Schmitt,
  Fabien Ringeval, Jing Han, Vedhas Pandit, Antoine Toisoul, Bjoern~W Schuller,
  et~al.
\newblock Sewa db: A rich database for audio-visual emotion and sentiment
  research in the wild.
\newblock {\em IEEE Transactions on Pattern Analysis and Machine Intelligence},
  2019.

\bibitem{Joo_2019_CVPR}
Hanbyul Joo, Tomas Simon, Mina Cikara, and Yaser Sheikh.
\newblock Towards social artificial intelligence: Nonverbal social signal
  prediction in a triadic interaction.
\newblock In {\em The IEEE Conference on Computer Vision and Pattern
  Recognition (CVPR)}, June 2019.

\bibitem{celiktutan2017multimodal}
Oya Celiktutan, Efstratios Skordos, and Hatice Gunes.
\newblock Multimodal human-human-robot interactions (mhhri) dataset for
  studying personality and engagement.
\newblock {\em IEEE Transactions on Affective Computing}, 2017.

\bibitem{alameda2015salsa}
Xavier Alameda-Pineda, Jacopo Staiano, Ramanathan Subramanian, Ligia Batrinca,
  Elisa Ricci, Bruno Lepri, Oswald Lanz, and Nicu Sebe.
\newblock Salsa: A novel dataset for multimodal group behavior analysis.
\newblock {\em IEEE Transactions on Pattern Analysis and Machine Intelligence},
  38(8):1707--1720, 2015.

\bibitem{poria2019meld}
Soujanya Poria, Devamanyu Hazarika, Navonil Majumder, Gautam Naik, Erik
  Cambria, and Rada Mihalcea.
\newblock Meld: A multimodal multi-party dataset for emotion recognition in
  conversations.
\newblock In {\em Proceedings of the 57th Annual Meeting of the Association for
  Computational Linguistics}, pages 527--536, 2019.

\bibitem{Rothbart:etal:2001}
Mary Rothbart, Stephan Ahadi, Karen Hershey, and Philip Fisher.
\newblock Investigations of temperament at three to seven years: The children's
  behavior questionnaire.
\newblock {\em Child development}, 72:1394--408, 09 2001.

\bibitem{Osa:2013}
Nuria Osa, Roser Granero, Eva Penelo, Josep Domènech, and Lourdes Ezpeleta.
\newblock The short and very short forms of the children's behavior
  questionnaire in a community sample of preschoolers.
\newblock {\em Assessment}, 21, 11 2013.

\bibitem{Ellis:Rothbart:2001}
Lesa Ellis and Mary Rothbart.
\newblock Revision of the early adolescent temperament questionnaire.
\newblock {\em Poster presented at the 2001 Biennal Meeting of the Society for
  Research in Child Development}, 01 2001.

\bibitem{Ashton:2009}
Michael Ashton and Kibeom Lee.
\newblock The hexaco-60: A short measure of the major dimensions of
  personality.
\newblock {\em Journal of personality assessment}, 91:340--5, 07 2009.

\bibitem{Gallardo-Pujol:2012}
David Gallardo-Pujol, Antonio Andres-Pueyo, and Alberto Maydeu-Olivares.
\newblock Maoa genotype, social exclusion and aggression: An experimental test
  of a gene-environment interaction.
\newblock {\em Genes, brain, and behavior}, 12, 10 2012.

\bibitem{williams2002investigations}
Kipling~D Williams, Cassandra~L Govan, Vanessa Croker, Daniel Tynan, Maggie
  Cruickshank, and Albert Lam.
\newblock Investigations into differences between social-and cyberostracism.
\newblock {\em Group dynamics: Theory, research, and practice}, 6(1):65, 2002.

\bibitem{abel2017cognitive}
Jennifer Abel and Molly Babel.
\newblock Cognitive load reduces perceived linguistic convergence between
  dyads.
\newblock {\em Language and Speech}, 60(3):479--502, 2017.

\bibitem{lindsay2017scoping}
Sally Lindsay, Kara~Grace Hounsell, and Celia Cassiani.
\newblock A scoping review of the role of lego{\textregistered} therapy for
  improving inclusion and social skills among children and youth with autism.
\newblock {\em Disability and health journal}, 10(2):173--182, 2017.

\bibitem{fusaroli2016heart}
Riccardo Fusaroli, Johanne~S Bj{\o}rndahl, Andreas Roepstorff, and Kristian
  Tyl{\'e}n.
\newblock A heart for interaction: Shared physiological dynamics and behavioral
  coordination in a collective, creative construction task.
\newblock {\em Journal of Experimental Psychology: Human Perception and
  Performance}, 42(9):1297, 2016.

\bibitem{hartup1993conflict}
Willard~W Hartup, Doran~C French, Brett Laursen, Mary~Kathleen Johnston, and
  John~R Ogawa.
\newblock Conflict and friendship relations in middle childhood: Behavior in a
  closed-field situation.
\newblock {\em Child Development}, 64(2):445--454, 1993.

\bibitem{underwood2004observational}
Marion~K Underwood, Bertrina~L Scott, Mikal~B Galperin, Gretchen~J Bjornstad,
  and Alicia~M Sexton.
\newblock An observational study of social exclusion under varied conditions:
  Gender and developmental differences.
\newblock {\em Child Development}, 75(5):1538--1555, 2004.

\bibitem{ajzen1987attitudes}
Icek Ajzen.
\newblock Attitudes, traits, and actions: Dispositional prediction of behavior
  in personality and social psychology.
\newblock In {\em Advances in experimental social psychology}, volume~20, pages
  1--63. Elsevier, 1987.

\bibitem{aschwanden2020cognitive}
Damaris Aschwanden, Mathias Allemand, and Patrick~L Hill.
\newblock Cognitive methods in personality research.
\newblock {\em The Wiley Encyclopedia of Personality and Individual
  Differences: Measurement and Assessment}, pages 49--54, 2020.

\bibitem{howard2017mobilenets}
Andrew~G. Howard, Menglong Zhu, Bo~Chen, Dmitry Kalenichenko, Weijun Wang,
  Tobias Weyand, Marco Andreetto, and Hartwig Adam.
\newblock Mobilenets: Efficient convolutional neural networks for mobile vision
  applications, 2017.

\bibitem{yang2016wider}
Shuo Yang, Ping Luo, Chen~Change Loy, and Xiaoou Tang.
\newblock Wider face: A face detection benchmark.
\newblock In {\em IEEE Conference on Computer Vision and Pattern Recognition
  (CVPR)}, 2016.

\bibitem{ghadiyaram2019large}
Deepti Ghadiyaram, Du~Tran, and Dhruv Mahajan.
\newblock Large-scale weakly-supervised pre-training for video action
  recognition.
\newblock In {\em IEEE Conference on Computer Vision and Pattern Recognition},
  pages 12046--12055, 2019.

\bibitem{tran2018closer}
Du~Tran, Heng Wang, Lorenzo Torresani, Jamie Ray, Yann LeCun, and Manohar
  Paluri.
\newblock A closer look at spatiotemporal convolutions for action recognition.
\newblock In {\em IEEE conference on Computer Vision and Pattern Recognition
  (CVPR)}, pages 6450--6459, 2018.

\bibitem{45611}
Shawn Hershey, Sourish Chaudhuri, Daniel P.~W. Ellis, Jort~F. Gemmeke, Aren
  Jansen, Channing Moore, Manoj Plakal, Devin Platt, Rif~A. Saurous, Bryan
  Seybold, Malcolm Slaney, Ron Weiss, and Kevin Wilson.
\newblock Cnn architectures for large-scale audio classification.
\newblock In {\em International Conference on Acoustics, Speech and Signal
  Processing (ICASSP)}. 2017.

\bibitem{DBLP:journals/corr/Abu-El-HaijaKLN16}
Sami Abu{-}El{-}Haija, Nisarg Kothari, Joonseok Lee, Paul Natsev, George
  Toderici, Balakrishnan Varadarajan, and Sudheendra Vijayanarasimhan.
\newblock Youtube-8m: {A} large-scale video classification benchmark.
\newblock {\em CoRR}, abs/1609.08675, 2016.

\bibitem{mensah2013global}
Yaw~M Mensah and Hsiao-Yin Chen.
\newblock Global clustering of countries by culture--an extension of the globe
  study.
\newblock {\em Available at SSRN 2189904}, 2013.

\bibitem{NIPS2017_7181}
Ashish Vaswani, Noam Shazeer, Niki Parmar, Jakob Uszkoreit, Llion Jones,
  Aidan~N Gomez, \L~ukasz Kaiser, and Illia Polosukhin.
\newblock Attention is all you need.
\newblock In I.~Guyon, U.~V. Luxburg, S.~Bengio, H.~Wallach, R.~Fergus,
  S.~Vishwanathan, and R.~Garnett, editors, {\em Advances in Neural Information
  Processing Systems 30}, pages 5998--6008. 2017.

\end{thebibliography}
}

\end{document}


\maketitle

\setlength{\parskip}{0.1in}
\setlength{\parindent}{15pt}

\renewcommand{\thesection}{S\arabic{section}}
\renewcommand{\thetable}{S\arabic{table}}
\renewcommand{\thefigure}{S\arabic{figure}}

\begin{table*}[t!]
\centering
	\setlength\tabcolsep{2.8pt}
	\scriptsize
\caption{Publicly available audiovisual human-human (face-to-face) dyadic interaction datasets. ``Interaction'', \textit{Acted} (actors improvising and/or following an interaction protocol, i.e. given topics/stimulus/tasks), \textit{Acted}$^*$ (Scripted), \textit{Non-acted} (natural interactions in lab environment) or \textit{Non-acted}$^*$ (non-acted but guided by interaction protocol); ``F/M'', number of participants per gender (Female/Male) or number of participants if gender is not informed; ``Sess'', number of sessions; ``Size'', hours of recordings;``\#Views'', number of RGB cameras used, and \textit{D} is RGB+D, \textit{E} is Ego, \textit{M} is Monochrome. The $\phi$ symbol is used to indicate missing/incomplete/unclear information on the source.}

\resizebox{\textwidth}{!}{%
\begin{tabular}{|C{1.95cm}|C{2.3cm}|C{1.4cm}|C{2cm}|C{3.5cm}|C{0.65cm}|C{0.8cm}|C{0.8cm}|C{0.9cm}|C{1cm}|}
\hline
\textbf{Name (Year)} 
& \textbf{Focus} 
& \textbf{Interaction} 
& \textbf{Modality} 
& \textbf{Annotations}
& \textbf{F/M}
& \textbf{Sess}
& \textbf{Size}
& \textbf{\#Views}
& \textbf{Lang.}\\ \hline

IFADV \cite{van2008ifadv} ('07)
& Speech \& conversation analysis 
& Non-acted
& Audiovisual
& Speech features, transcripts
& 24/10
& 20
& 5h 
& 2
& Dutch \\ \hline

IEMOCAP \cite{Busso2008} ('08)
& Emotion recognition 
& Acted$^*$ \& Acted 
& Audiovisual, face \& hands MoCap.
& Emotions, transcripts, turn-taking
& 5/5
& 5
& $\sim$12h 
& 2
& English \\ \hline

CID \cite{blache2017corpus} ('08)
& Speech \& conversation analysis
& Non-acted \& Non-acted$^*$
& Audiovisual
& Speech features, transcripts
& 10/6
& 8
& 8h 
& 1
& French \\ \hline

Spontal \cite{edlund2010spontal} ('10)
& Speech \& conversation analysis 
& Non-acted \& Non-acted$^*$
& Audiovisual, head \& torso MoCap.
& Transcripts, speech features
& $\phi$ 
& 120
& 60h 
& 2
& Swedish \\ \hline

NOMCO \cite{paggio2010nomco} ('10)
& Speech \& conversation analysis 
& Non-acted \& Non-acted$^*$
& Audiovisual
& Speech \& interaction features, gestures, transcripts, emotions
& 6/6 $^\phi$
& 60
& $\sim$6h 
& 3
& Danish, Swedish, Finnish \\ \hline

HUMAINE$^{\dag}$ \cite{douglas2007humaine,douglas2011humaine} ('11)
& Emotion analysis
& Non-acted$^*$
& Audiovisual
& Emotions
& 34
& 18
& $\sim$12h 
& 4
& English \\ \hline

MMDB \cite{rehg2013decoding} ('13) 
& Adult-infant interaction analysis
& Non-acted$^*$
& Audiovisual, depth, physiological
& Social cues (gaze, vocal affects, gestures...)
& 121
& 160
&$\sim$13.3h
& 8 + 1D
& English \\ \hline

MAMCO \cite{vella2013overlaps} ('14)
& Overlap analysis
& Non-acted
& Audiovisual
& Transcripts
& 6/6
& 12
&$\sim$1h 
& 3
& Maltese \\ \hline

4D CCDb \cite{marshall20154d} ('15)
& Speech \& conversation analysis
& Non-acted
& Audiovisual, depth
& Facial expressions, head gestures, utterances
& 2/2
& 17
& $\sim$0.2h
& 6 + 8M
& English \\ \hline

MAHNOB \cite{bilakhia2015mahnob} ('15)
& Mimicry
& Non-acted$^*$
& Audiovisual, head MoCap.
& Head, face and hand gestures, personality scores (self-reported)
& 29/31
& 54
& 11.6h
& 2 + 13M
& English \\ \hline

MIT Interview \cite{naim2015automated} ('15) 
& Hirability analysis 
& Non-acted$^*$
& Audiovisual
& Hirability, speech features, social \& behavioral traits, transcripts
& 43/26
& 138
& 10.5h
& 2
& English \\ \hline

MPIIEMO \cite{muller2015emotion} ('15)
& Bodily emotion analysis
& Acted
& Audiovisual
& Emotions
& 3/2
& 8$\times$7$\times$4 (tasks)
& $\sim$2.4h
& 8
& German$^\phi$ \\ \hline 

JESTKOD \cite{bozkurt2015jestkod} ('15)
& Agreement classification
& Non-acted$^*$
& Audiovisual, body MoCap.
& Agreement, emotion
& 4/6
& 25
& 4.3h
& 1
& Turkish \\ \hline

Creative IT \cite{Metallinou:2016} ('16)
& Emotion recognition
& Acted
& Audiovisual, body MoCap.
& Transcripts, speech features, emotion
& 9/7
& 8
& $\sim$1h
& 2
& English \\ \hline

MSP-IMPROV \cite{Busso_2017} ('17)
& Emotion recognition
& Acted \& Non-acted
& Audiovisual
& Turn-taking, emotion
& 6/6
& 6
& 9h
& 2
& English \\ \hline

NNIME \cite{chou2017nnime} ('17) 
& Emotion analysis
& Non-acted$^*$
& Audiovisual, physiological
& Emotion, transcripts
& 22/20
& 102
& $\sim$11h
& 1
& Chinese \\ \hline

RAMAS \cite{perepelkina2018ramas} ('18) 
& Emotion analysis 
& Non-acted$^*$ \& Acted
& Audiovisual, depth, body MoCap.
& Physiological signals, emotion, interaction traits 
& 5/5
& 80 
& $\sim$7h 
& 2 + 1D
& Russian \\ \hline

DAMI-P2C \cite{chen2020dyadic} ('20)
& Adult-infant interaction analysis
& Non-acted$^*$ 
& Audiovisual
& Emotion, sociodemographics, parenting assessment, child personality (peer-reported)
& 38/30
& 65
& $\sim$21.6h
& 1 $^\phi$
& English \\ \hline

\textbf{UDIVA} (ours) ('20) 
& Social interaction analysis
& $\frac{1}{5}$ Non-acted \& $\frac{4}{5}$Non-acted$^*$ 
& Audiovisual, heart rate 
& Personality scores (self- \& peer- reported), sociodemographics, mood, fatigue, relationship type
& 66/81
& 188$\times$5 (tasks)
& 90.5h
& 6 + 2E
& Spanish, Catalan, English \\ \hline
\multicolumn{9}{l}{$^{\dag}$ Here we consider the Green Persuasive and the EmoTABOO \cite{zara2007collection} datasets together.}\\

\end{tabular}%
}
\label{tab:databases}
\end{table*}

Here we provide further information on some sections of the paper. First, we include an extended table containing a more complete comparison of available dyadic interaction datasets. Then, we describe the rationale behind using 32 frames per video chunk and the procedure used to crop the face-only videos, as part of the proposed methodology. Finally, we detail the training strategy, the algorithm used to define the data splits (such that a balance was kept on the participants and sessions features) and report the resulting distribution of OCEAN values among them.

\section{Face-to-face dyadic datasets comparison (Sec.~2)}
For the sake of completeness, Table~\ref{tab:databases} contains an extended review of publicly available face-to-face dyadic interactions datasets that contain at least audiovisual data. Most of the datasets are tailor-made for too specific purposes or limited in the number of participants, recordings, views, context annotations or language. Hence, there is no big enough general purpose database in the literature that could allow for an integral analysis of both, the interaction and the participants.

\section{Size of video chunks (Sec.~4.1)}
The original Video Action Transformer~\cite{Girdhar_2019_CVPR} uses an I3D backbone pretrained on Kinetics-400~\cite{carreira2017quo} for spatiotemporal feature extraction. Such backbone uses 64 frames per chunk, which is equivalent to around 3 seconds of video. Instead, we opted for the R(2+1)D backbone~\cite{tran2018closer} pretrained on IG-65M dataset, which has shown to provide significant performance gains~\cite{ghadiyaram2019large}. This backbone uses 32 frames per chunk, so by using a stride of 2 we manage to encode approximately the same time window as the original method with half the number of frames while reducing the memory load. This is equivalent to downsampling the original videos from 25 fps to 12.5, that is, 1 frame every 0.08 seconds. Although not frequent, there is a chance to miss some fast-paced facial and body micro-actions in such downsampling process. However, there is also the trade-off we try to balance between losing some of these fast micro-actions and being able to include a larger, and also important, temporal context.

\section{Face detection and tracking (Sec.~4.1)}
As described in the main paper, we use a face chunk video as one of the inputs of the model, which is used together with the participants' metadata to form the query of the transformer model. In order to detect the faces we use MobileNet-SSD~\cite{howard2017mobilenets}, deployed using Tensorflow Object Detection API \cite{8099834} and pretrained on the Wider Face Dataset~\cite{yang2016wider}. As we consider only frontal cameras, the detection task is not very challenging, therefore, on more than 95\% of the videos the detection ratio is higher than 75\%. In case the gap between consecutive detections is lower than 25 frames (1 second), we linearly interpolate the coordinates of the boxes. Since there are frames in which the frontal cameras capture both participants, we need to identify the target person before computing the face chunks. In order to do so, we employ a basic tracking algorithm based on the following 2 steps: (1) \textit{identify} target person's face: given a video, the face of the target person is considered the first detection that has a mean intersection over union (IoU) score higher than 0.2 with respect to all the other faces in the video; (2) \textit{track} target person face throughout the video based on the IoU.

\begin{figure*}[t!]
\centering    
\includegraphics[width=\textwidth]{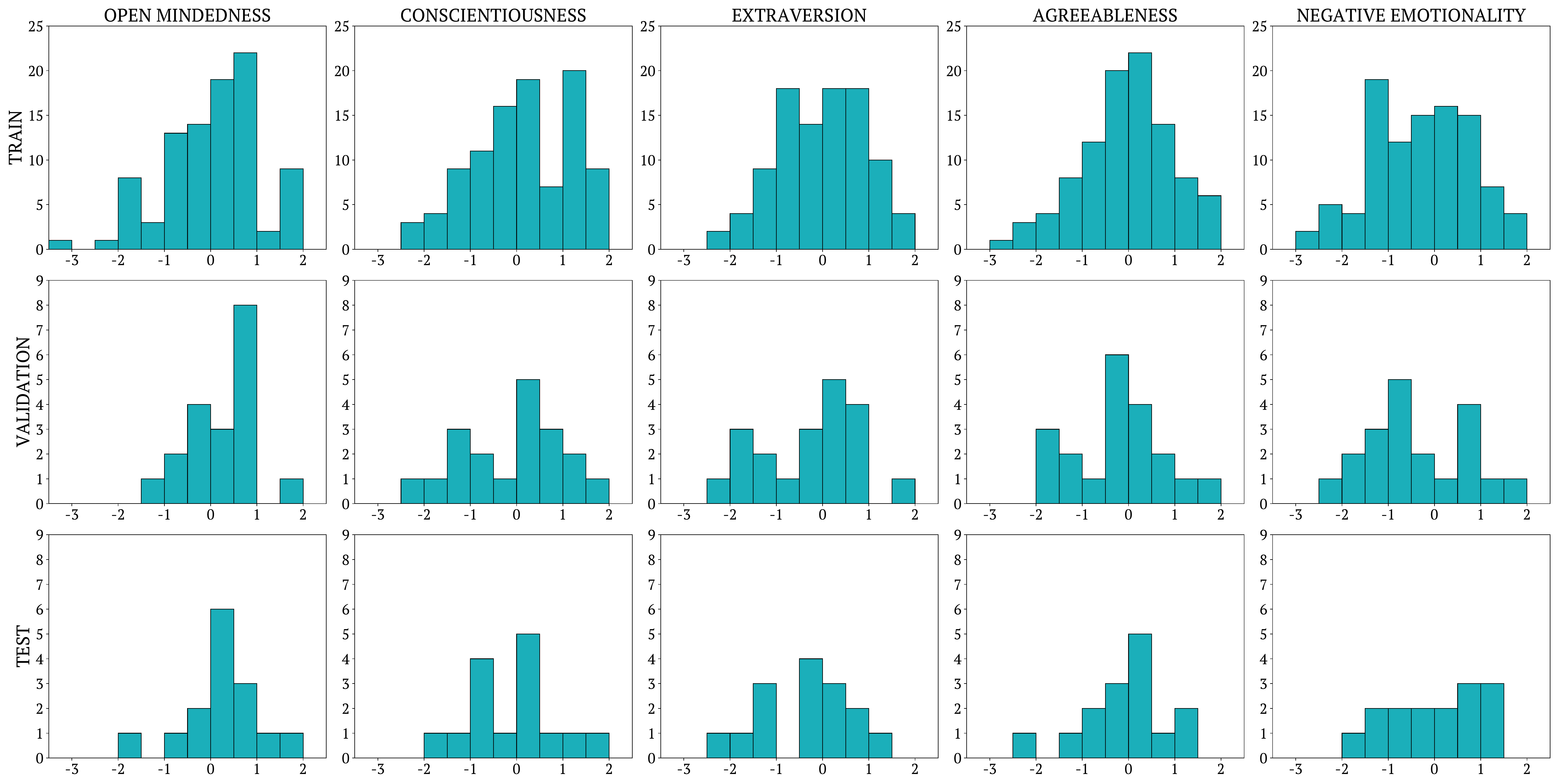}
\caption{Distribution of the self-reported personality trait (OCEAN) values across train, validation and test splits used to evaluate the proposed personality inference method. X axis refers to z scores for each personality trait. Y axis refers to number of participants.}
\label{fig:ocean:dist:subset}
\end{figure*}

\section{Training strategy (Sec.~4.2)} 
\vspace{-0.2cm}
The proposed model was trained using Adam optimizer with $\beta_1 = 0.9$, $\beta_2 = 0.999$, $\epsilon = 1e-8$ and a learning rate of $1e-5$. We used a batch size of 2 and the Mean Squared Error as the loss function.
We compute the validation error approximately 30 times per epoch and select the model that gives the best results considering the mean with its previous and next evaluation scores. The final results, detailed in Sec.~4.3 of the main paper, were obtained by freezing the layers of the R(2+1)D backbones, as strategies such as finetuning end-to-end or only the last block of the feature extractors led to fast overfitting. 
\vspace{2cm}

\section{Personality trait (OCEAN) values over splits (Sec.~4.2)}
In this section, we briefly describe the procedure used to define the data splits used during the experiments described in the experimental section.

In order to split the data among training, validation and test subsets, some sessions needed to be removed so that no participants were repeated in any of the subsets. The final split was selected using a greedy optimization method that iteratively removed and added sessions based on their importance until a valid split ratio was found. Such importance was determined by the groups distribution and the number of remaining sessions per participant. In particular, the method tried to minimize a set of costs to: (1) ensure that distributions among splits were not different by means of a Kolmogorov-Smirnov significance test~\cite{massey1951kolmogorov}; (2) ensure that Pearson's correlation of gender, age and personality values among splits did not differ by a large margin; (3) attempt to have a uniform distribution in validation and test with respect to age and gender to correct selection bias; (4) attempt to have a close-to-uniform distribution of group combinations; and (5) try to maximize the number of sessions without losing participants, while considering also the train/validation/test ratio. The resulting distribution of OCEAN values among splits can be seen in Fig.~\ref{fig:ocean:dist:subset}.

{\small

}